%% file: main.tex
\documentclass{article}
\usepackage{amsfonts} 
\usepackage{amsmath} 
\usepackage{url} 
\usepackage{booktabs} 
\usepackage[utf8]{inputenc}
\usepackage{graphicx}
\usepackage{float}
\usepackage{wrapfig}
\restylefloat{figure}
\usepackage[ruled]{algorithm2e} 
\graphicspath{{img/}}
\usepackage{subcaption}
\usepackage{tikz}
\usepackage{placeins}
\usepackage[export]{adjustbox}
\usepackage{booktabs} 
\usepackage{authblk}

\input{dropshadow}

\newcommand{\added}[1]{{#1}}

\begin{document}

\title{A new approach for pedestrian density estimation using moving sensors and computer vision}


\author[1]{Eric K. Tokuda}
\author[3]{Yitzchak Lockerman}
\author[1]{Gabriel B. A. Ferreira}
\author[2]{Ethan Sorrelgreen}
\author[2]{David Boyle}
\author[1]{Roberto M. Cesar-Jr.}
\author[3]{Claudio T. Silva}

\affil[1]{University of São Paulo, Brazil} 
\affil[2]{Carmera Inc., USA} 
\affil[3]{New York University, USA} 
\maketitle

\begin{abstract}
An understanding of person dynamics is indispensable for numerous urban applications including the design of transportation networks and planing for business development. Pedestrian counting often requires utilizing manual or technical means to count individuals in each location of interest. However, such methods do not scale to the size of a city and a new approach to fill this gap is here proposed. In this project, we used a large dense dataset of images of New York City along with computer vision techniques to construct a spatio-temporal map of relative person density. Due to the limitations of state of the art computer vision methods, such automatic detection of person is inherently subject to errors. We model these errors as a probabilistic process, for which we provide theoretical analysis and thorough numerical simulations. We demonstrate that, within our assumptions, our methodology can supply a reasonable estimate of person densities and provide theoretical bounds for the resulting error. 
\end{abstract}




\input{introduction}

\input{method}

\input{experiments}
\input{conclusion}
\input{ack}

\bibliographystyle{plain}
\bibliography{main}

\input{appendix}
\end{document}

%% file: dropshadow.tex
\usetikzlibrary{shadows,calc}

\def\shadowshift{3pt,-3pt}
\def\shadowradius{6pt}

\colorlet{innercolor}{black!60}
\colorlet{outercolor}{gray!05}

\newcommand\drawshadow[1]{
    \begin{pgfonlayer}{shadow}
        \shade[outercolor,inner color=innercolor,outer color=outercolor] ($(#1.south west)+(\shadowshift)+(\shadowradius/2,\shadowradius/2)$) circle (\shadowradius);
        \shade[outercolor,inner color=innercolor,outer color=outercolor] ($(#1.north west)+(\shadowshift)+(\shadowradius/2,-\shadowradius/2)$) circle (\shadowradius);
        \shade[outercolor,inner color=innercolor,outer color=outercolor] ($(#1.south east)+(\shadowshift)+(-\shadowradius/2,\shadowradius/2)$) circle (\shadowradius);
        \shade[outercolor,inner color=innercolor,outer color=outercolor] ($(#1.north east)+(\shadowshift)+(-\shadowradius/2,-\shadowradius/2)$) circle (\shadowradius);
        \shade[top color=innercolor,bottom color=outercolor] ($(#1.south west)+(\shadowshift)+(\shadowradius/2,-\shadowradius/2)$) rectangle ($(#1.south east)+(\shadowshift)+(-\shadowradius/2,\shadowradius/2)$);
        \shade[left color=innercolor,right color=outercolor] ($(#1.south east)+(\shadowshift)+(-\shadowradius/2,\shadowradius/2)$) rectangle ($(#1.north east)+(\shadowshift)+(\shadowradius/2,-\shadowradius/2)$);
        \shade[bottom color=innercolor,top color=outercolor] ($(#1.north west)+(\shadowshift)+(\shadowradius/2,-\shadowradius/2)$) rectangle ($(#1.north east)+(\shadowshift)+(-\shadowradius/2,\shadowradius/2)$);
        \shade[outercolor,right color=innercolor,left color=outercolor] ($(#1.south west)+(\shadowshift)+(-\shadowradius/2,\shadowradius/2)$) rectangle ($(#1.north west)+(\shadowshift)+(\shadowradius/2,-\shadowradius/2)$);
        \filldraw ($(#1.south west)+(\shadowshift)+(\shadowradius/2,\shadowradius/2)$) rectangle ($(#1.north east)+(\shadowshift)-(\shadowradius/2,\shadowradius/2)$);
    \end{pgfonlayer}
}

\pgfdeclarelayer{shadow} 
\pgfsetlayers{shadow,main}

\newsavebox\mybox
\newlength\mylen

\newcommand\shadowimage[2][]{%
\setbox0=\hbox{\includegraphics[#1]{#2}}
\setlength\mylen{\wd0}
\ifnum\mylen<\ht0
\setlength\mylen{\ht0}
\fi
\divide \mylen by 120
\def\shadowshift{\mylen,-\mylen}
\def\shadowradius{\the\dimexpr\mylen+\mylen+\mylen\relax}
\begin{tikzpicture}
\node[anchor=south west,inner sep=0] (image) at (0,0) {\includegraphics[#1]{#2}};
\drawshadow{image}
\end{tikzpicture}}

%% file: introduction.tex
\section{Introduction}

\begin{figure}
        \centering
        \shadowimage[width=0.27\textwidth]{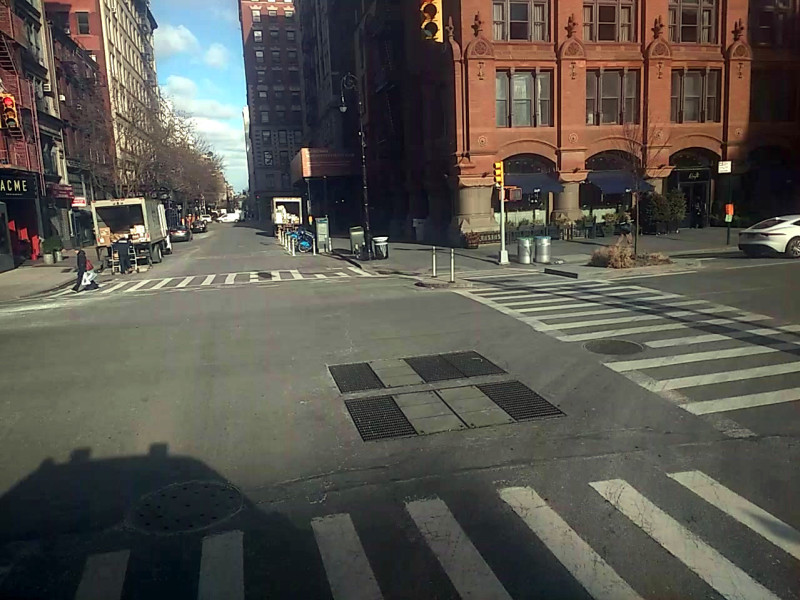} 
        \shadowimage[width=0.27\textwidth]{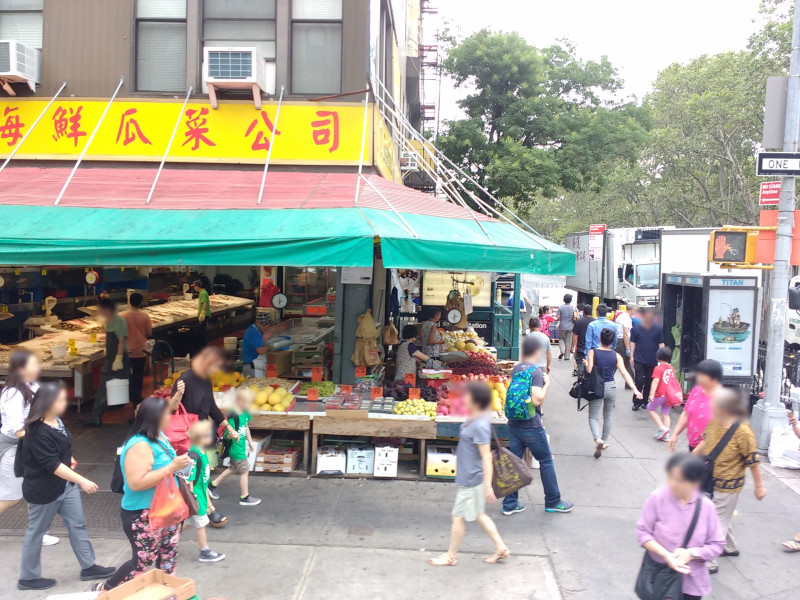} 
        \shadowimage[width=0.27\textwidth]{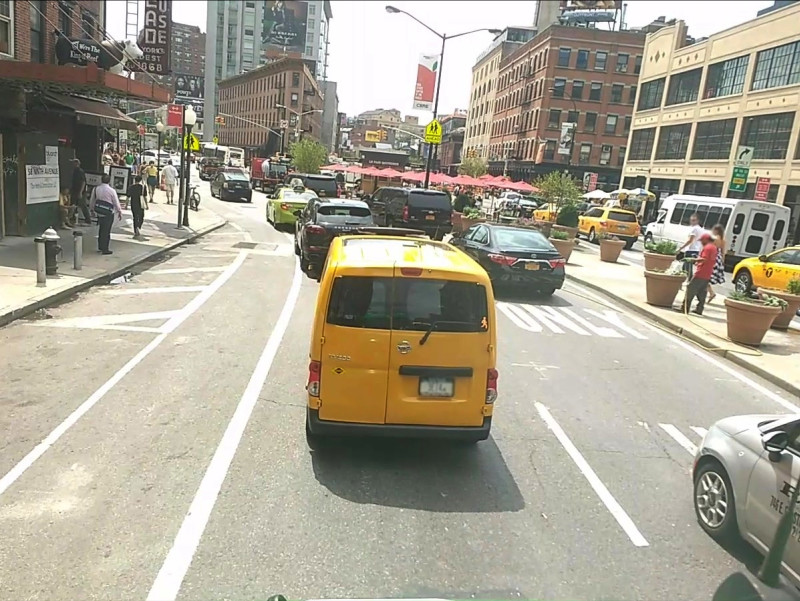} 
        \shadowimage[width=0.27\textwidth]{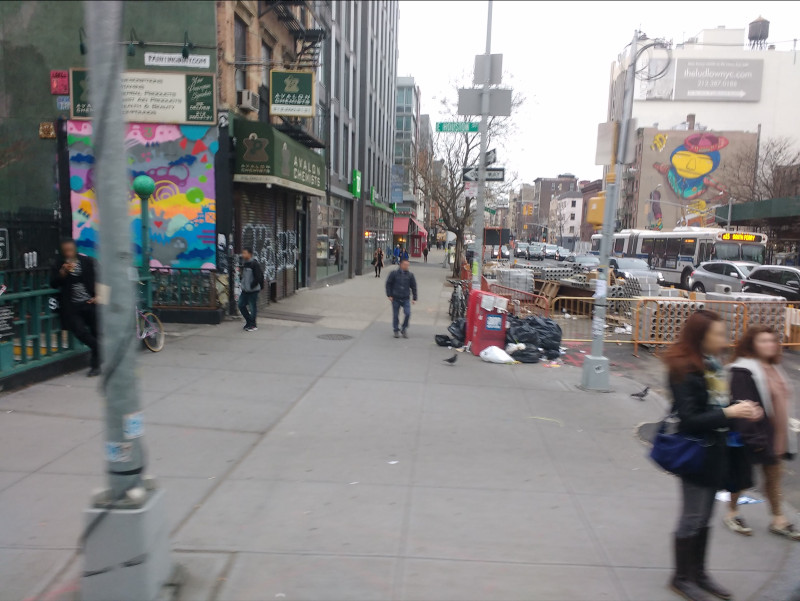} 
        \shadowimage[width=0.27\textwidth]{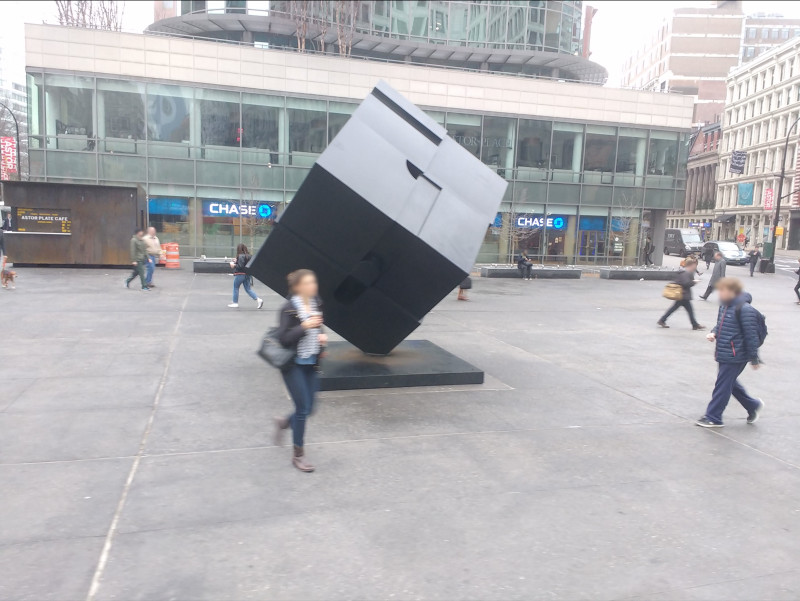} 
        \shadowimage[width=0.27\textwidth]{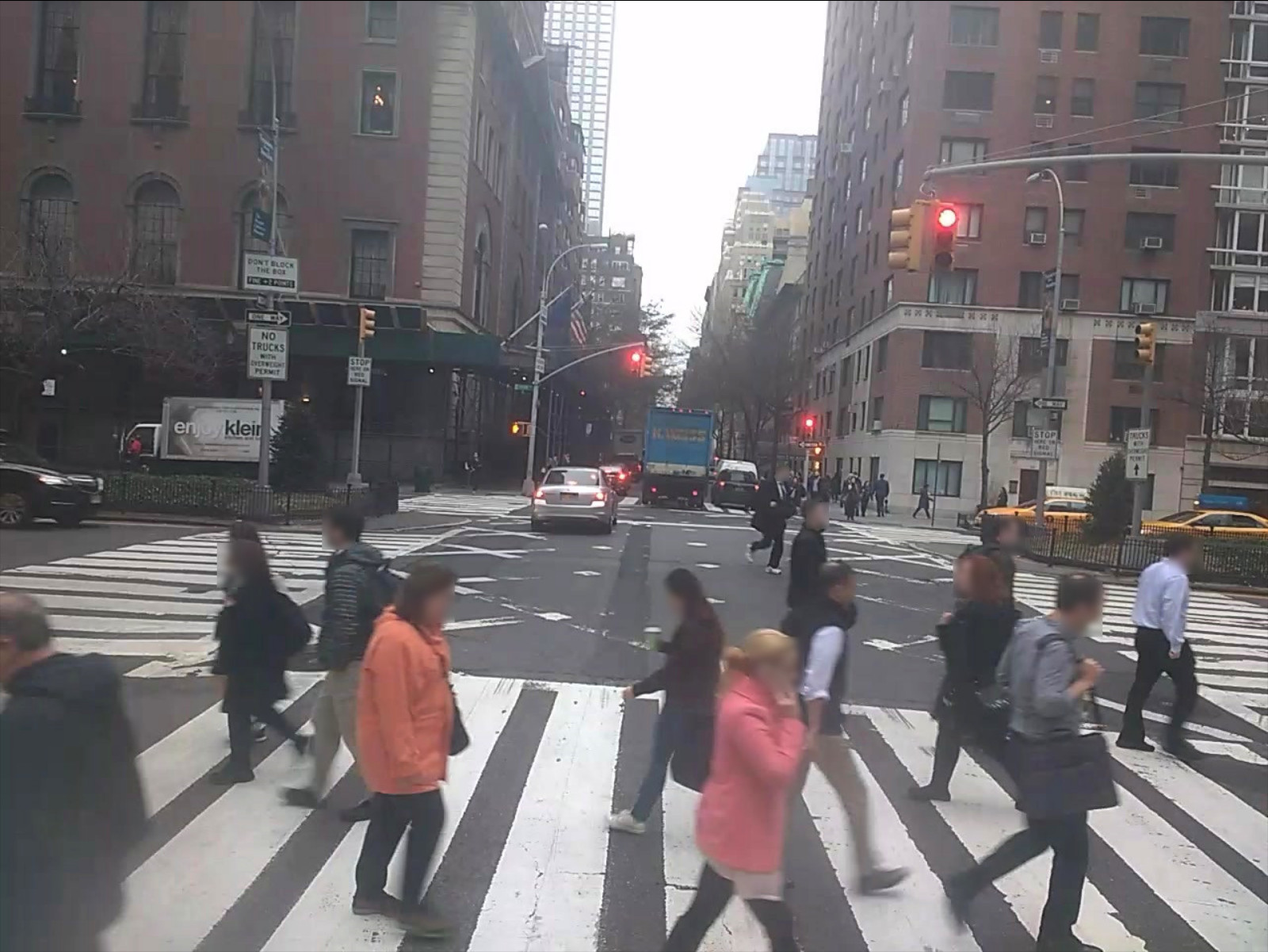} 
	\caption{Sample of the images provided by Carmera. \added{More than 10 million images have been used for the people density estimation evaluated in this work}.}
        \label{fig:sample}
\end{figure}

Pedestrians represent an integral and pervasive aspect of the urban environment. Real estate, consumer patterns, public safety, and other aspects of city life are deeply intertwined with the variations of distribution of people across a city. However, current methods for estimating the distribution of people within a city tend to be expensive and mostly produce a sparse sampling of a few locations. 


 In this paper we propose a novel method to estimate the pedestrians distribution using street-level pictures of the city. We determine bounds for the estimate and apply the methodology using a large and rich collection of street-level pictures from New York city to obtain a dense pedestrian map of the region.

We make use of state-of-art computer vision methods to identify and count pedestrians and we take into account the errors inherent to the detection process, we model it as a \emph{probabilistic detection} and provide a closed form and bounds for the asymptotic error of the sampling process. We compare these formulas to numerical simulations of the sensing process. Our results suggests that computer vision produces usable data, despite the inherent noise.  We utilized over 40 million street-level images provided by Carmera~\footnote{http://www.carmera.com}. These images were obtained via their partnerships with high coverage fleets operating daily on city streets that traveled through the region of Manhattan Island in New York City over the course of a year. A deep-learning based algorithm for pedestrian detection was utilized in a case study to map the density of pedestrians in Manhattan.

The contributions of this paper can be summarized as 
\begin{enumerate}
  \item A new method for the analysis of the distribution of people using state of the art, but imperfect, computer vision algorithms.
  \item A closed form function and bounds for the asymptotic error of the resulting person densities.
  \item The results of simulations validating the sampling process and the derived asymptotic error.
  \item A case study demonstrating the resulting densities for a collection of images from the City of New York. 
\end{enumerate}



\subsection{Related Work} 

There are many ongoing efforts on the use of urban data to achieve citizen-centered improvements~\cite{zheng2014urban}. Governments and organizations in urban environments collect a vast amount of data daily~\cite{united2016air} encompassing a large assortment of information including mobility, crime and pollution. The collection and use of this information has been attracting attention from the academics, governments and corporations \cite{vanegas2012automatic}. Arietta et al.~\cite{arietta2014city} explores the correlation of visual appearance of pictures and the attributes of the region it pertains. They collected images from Google Inc.~\cite{googlemaps} and also indicators from multiple regions and trained a model~\cite{burges1998tutorial} to predict the indicator based on images. The city attributes include violent crime rates, theft rates, housing prices, population density and trees presence. Results show that the visual data can be efficiently used to predict the region attributes. Additionally, the regressor trained in one region showed reasonable results when tested in a different city. \added{In a different problem, Zemene et al.~\cite{zemene2018large} proposes a technique of determining the geo-location of the scene depicted in an image. Local and global features are extracted and used in a posterior features matching step. They report high accuracy and speed performances compared to existing approaches.}

A citywide count of people has numerous applications for urban planners including the design of public transport network and of public spaces~\cite{whyte2012city}. One approach is to have people scattered around the city manually counting the people nearby. However, this approach is laborious because it requires dedicated people to perform the measurements. Another possibility explored  in Reades et al.~\cite{reades2007cellular} is to use cellphone use data to perform the person count. One clear limitation of this approach is that these data are not public and their coverage are restricted to the places where the carrier signal is present. Additionally, it is hard to know whether the cellphone signal is from someone in a building or in a car.

Alternatively, we can consider the visual task of finding people in city images. A remarkable work in this task consists in using the histogram of oriented gradients as the features vector and a support vector machines for the classification task~\cite{dalal2005histograms}. In the context of deep neural networks~\cite{krizhevsky2012imagenet,szegedy2015going}, the work of Ren et al.~\cite{ren2015faster} introduced an approach that tries to solve this task by using a unified network that performs  region proposal and  classification. In this way, the method accepts annotations of multiple sized objects during the training step and during the testing stage, it performs classification of those objects in images of arbitrary sizes. In Dai et al.~\cite{dai16rfcn} the authors follow the two-stage region proposal and classification framework of Ren et al.~\cite{ren2015faster} and proposes the Region-based Fully Convolutional Networks (R-FCN) which incorporate the idea of position-sensitive score maps to reduce the computational burden by sharing the per-RoI computation. Such speed alterations allow the incorporation of classification backbones such as Ren et al.~\cite{he2016deep}. \added{Alternatively to the traditional person detection, head detection~\cite{gao2016people,saqib2018person} approaches have also been explored for efficient person count estimates. Given that information from the body is disregarded, however, it is natural to expect an inferior accuracy in complex environments.}

There are several city images repositories that contemplate people, some of them obtained using static cameras~\cite{vezzani2010video,oh2011large} and others obtained using dynamic ones~\cite{geiger2013vision,cordts2016cityscapes,maddern20171}. Such configuration of sensors arrangement have long been studied in the \emph{sensor network} field~\cite{akyildiz2007survey,akyildiz2002survey,othman2012wireless} and an important aspect of these networks is whether the sensors are static or mobile. In Wang et al.~\cite{wang2003bidding} the authors explore the setting of a network composed of both static sensors and of mobile sensors. The holes in the coverage of the static sensors network are identified and the mobile sensors are used to cover the holes. A common problem in sensor networks is the \emph{k-coverage problem} defined in Huang et al.~\cite{huang2005coverage}, that aims to find the optimal setting of sensors such that any region is covered at least by k sensors.  In Yang et al~\cite{yang2003counting} the authors perform the task of counting people based on images obtained through a wireless network of static sensors. 

Apart from controllable mobile sensors network, many works explore data collected from collaborative uncontrolled sensors~\cite{basagni2007controlled} such as from vehicles GPS~\cite{shi2009automatic,karagiorgou2017layered}, mobile phones  sensors~\cite{sheng2012energy,lane2010survey,rana2010ear} and even from on-body sensors~\cite{consolvo2008activity}.


The work of Li et al.~\cite{li2017city} considers the problem of using GPS data from a network of uncontrolled sensors to reconstruct the traffic in a city. They do that in two steps: initial traffic reconstruction and dynamic data completion. Such approach allowed the authors to get a complete traffic map and a 2D visualization of the traffic.

There are many ways to model the movement of mobile nodes in a sensor network, the so-called \emph{mobility models}~\cite{camp2002survey}. A simple one is the \emph{random walk mobility model}~\cite{davies2000evaluating} where at each instant in time each particles gets  a direction and a speed to move. In the \emph{random waypoint mobility model}~\cite{johnson1996dynamic}, in turn, particles are given destinies and speeds. They travel toward their goal and once they get the destination a new goal and speed are given. The \emph{Gauss-Markov mobility model}~\cite{liang1999predictive} attempts to eliminate abrupt stops and sharp turns present in the random waypoint mobility model. It is done by computing the current position based on the previous position, speed and direction.
 

Simulation of wireless sensor networks has long been studied~\cite{vinyals2011survey, lesser2012distributed, niazi2011novel} because it allows a complete analysis of system architectures by providing a controlled environment for the system~\cite{titzer2005avrora}. The real-life systems non-determinism is simulated by the use of pseudo random number generators~\cite{knuth1997art}. Among the large number of pseudo random number generators~\cite{park1988random}, a popular algorithm is the Mersenne Twister~\cite{matsumoto1998mersenne} due to its efficiency and robustness.

%% file: method.tex
\section{Data and proposed method}
\subsection{Data}
The images utilized in this work come from 
a fleet of camera equipped cars (similar to Lee et al.~\cite{lee2009dissemination}) traveling through Manhattan. Carmera provides a temporally and spatially dense collection of pictures. \added{The volume of the sample utilized is at least an order of magnitude higher compared to the most popular images datasets (see Table~\ref{tab:datasets}). It also differs from urban image databases~\cite{googlemaps,vezzani2010video,oh2011large,geiger2013vision,cordts2016cityscapes,maddern20171} by providing  dense temporal coverage in addition to dense spatial coverage.} The orientation of the cameras varies and the nature of the images are similar to street level collections provided by many mapping services. However, the images are not stitched into a $360$ degree panorama. Every image is accompanied by metadata including the acquisition time, location, and camera orientation. The images are captured as the vehicle travels, with no control of the content, the illumination, the weather, the traffic conditions, or vehicular speed. The typical image depicts a urban scenario as a background and the city dynamics including person, vehicles and bicycles such as in Figure~\ref{fig:sample}.

\begin{table}[ht!]
	\caption{\added{Comparison of the numbers of images utilized to other popular datasets.}}
	\centering

{
	\begin{tabular}{lcc}
	\toprule
		Datasets & Number of images & Resolution\\ \midrule
		CIFAR 10 \cite{krizhevsky2009learning}&  $6.0~\cdot 10^3$ & low\\
		Pascal VOC 2007~\cite{everingham2010pascal}&  $9.9 \cdot 10^3$ & middle\\
		Cityscapes \cite{cordts2016cityscapes}&  $2.0 \cdot 10^4$ & very high\\
		MS COCO \cite{lin2014microsoft}&  $3.3 \cdot 10^5$ & high\\
		ILSVRC 2017 \cite{russakovsky2015imagenet}&  $1.2 \cdot 10^6$ & high\\
		\textbf{Our sample} &  \textbf{$1.0 \cdot 10^7$} & \textbf{high}\\
	\bottomrule
	\label{tab:datasets}
	\end{tabular}
}
\end{table}

\begin{figure}
    \centering
	\includegraphics[width=.45\textwidth]{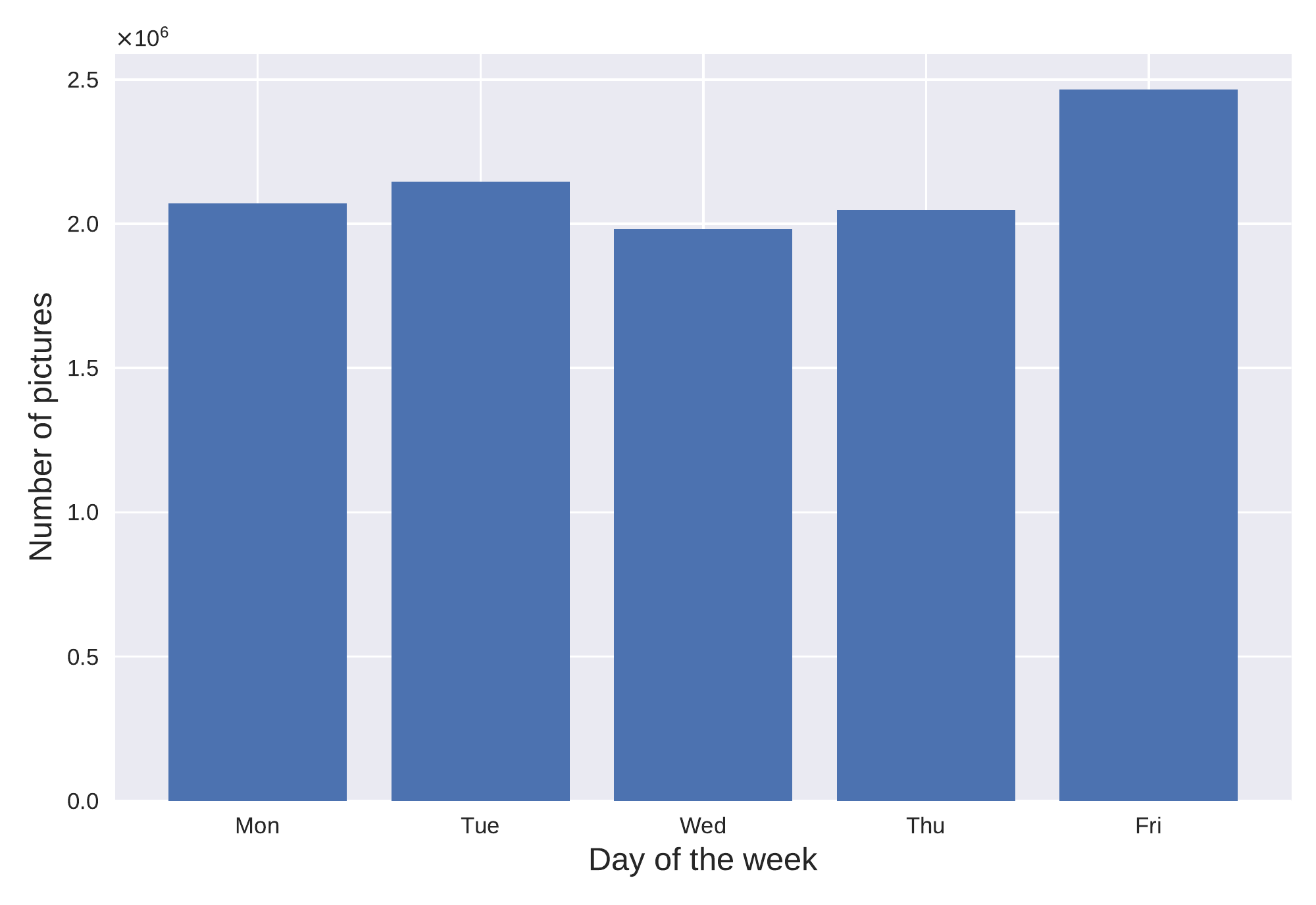}
	\qquad
	\includegraphics[width=.45\textwidth]{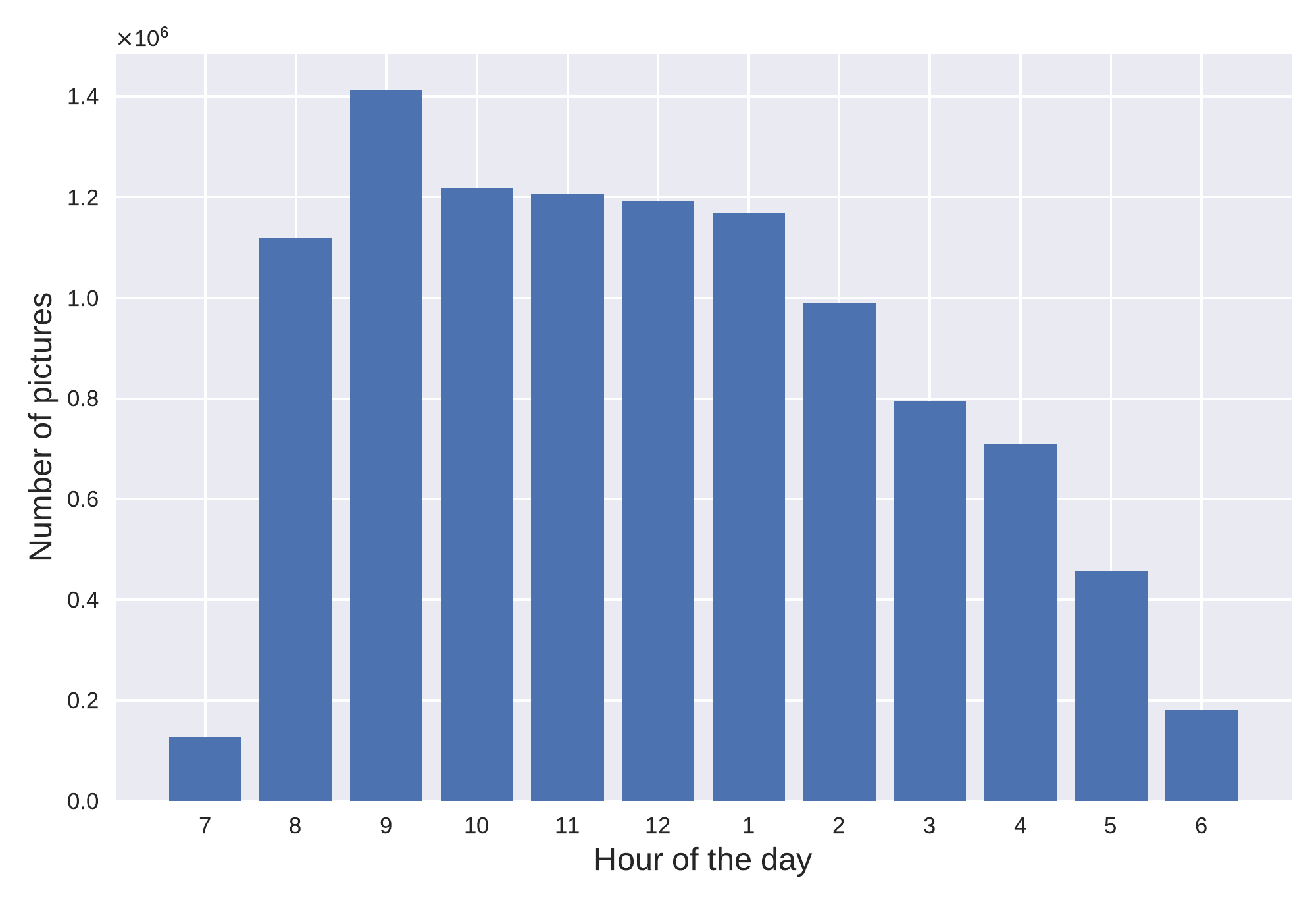}
	\caption{\added{Temporal distribution of the collection of images by day of the week (left) and by hour of the day (right).}}
        \label{fig:timeanalysis}
\end{figure}

\begin{figure}
    \centering
	\includegraphics[width=.6\textwidth]{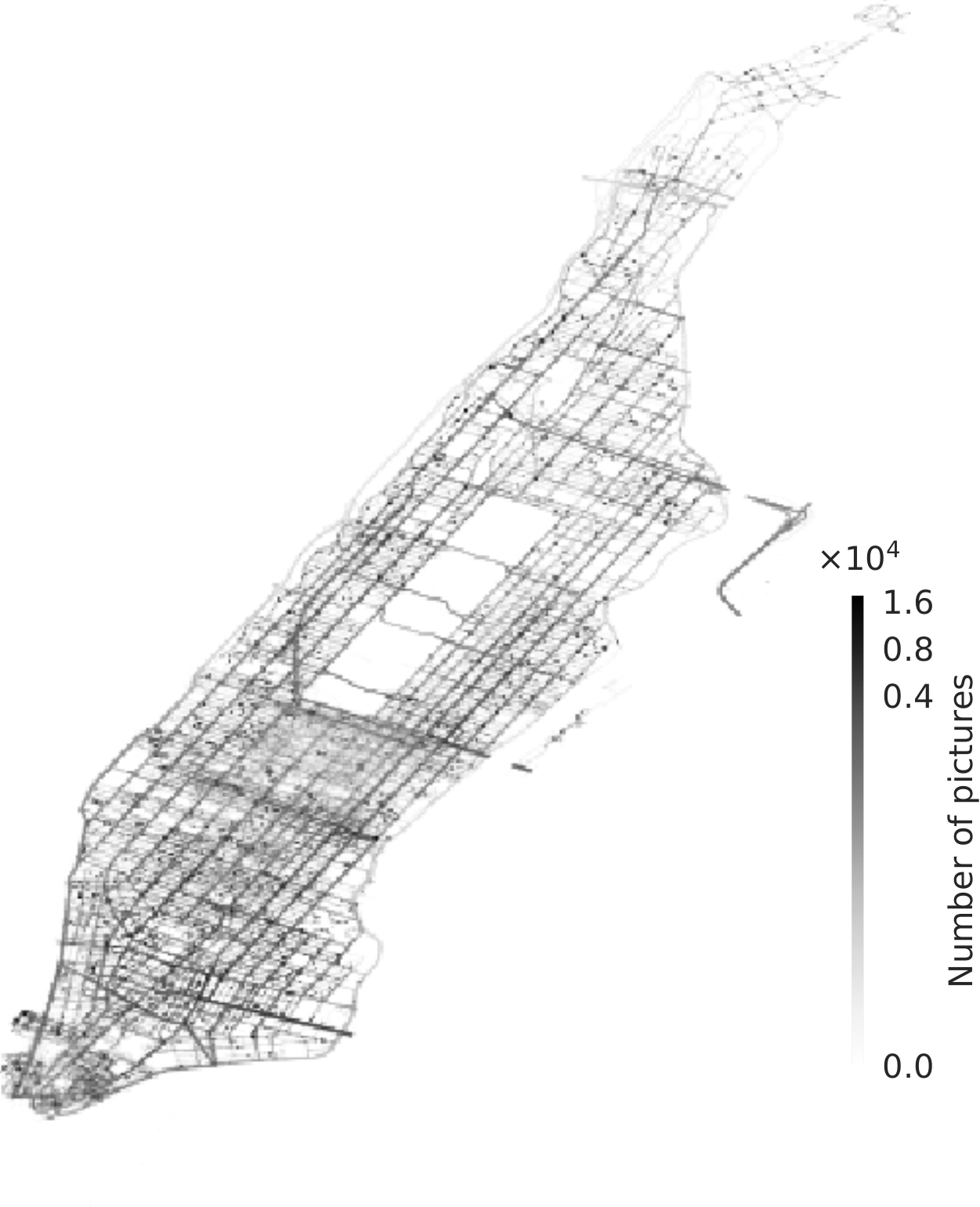}
	\caption{\added{Geographical distribution of the number of pictures utilized in the region of Manhattan. The highest concentration of images are in the bottom and middle regions of map.}}
        \label{fig:geoanalysis}
\end{figure}

All images included in the sample have a resolution of $1280\times960$. 
We used a sample of images captured from March 2016 to February 2017 containing  10,708,953 images. This sample presents a dense spatial sampling of the whole region \added{of Manhattan (see Figure~\ref{fig:geoanalysis})} over a year and irregular spatio-temporal sampling on a daily basis (see Figure~\ref{fig:timeanalysis}) All resulting heatmaps are weighted sampling according to this distribution. \added{Despite the large scale of the collection utilized in this work, it represents a small portion of the full dataset obtained by the company.}

\label{subsec:cv}

\subsection{Person detection}
We adopt the same metric as Everingham et al. \cite{everingham2010pascal} when comparing the detected objects in an image to the ground-truth. A detected object is considered to correspond to a particular ground truth object if there is a minimum ratio of $50\%$ between the overlap of the detected bounding boxes $B_{detected}$ ground-truth bounding boxes $B_{gtruth}$, and the union of the two areas (see Equation~\ref{eq:iou}). 

\begin{equation}
\frac{\left|B_{detected} \cap B_{gtruth}\right|}{\left|B_{detected} \cup B_{gtruth}\right|} >= 0.5
\label{eq:iou}
\end{equation}

The recognition of distant objects in an image is difficult for humans and is even more difficult for computers. We assume that, on average, the size of a person within an image is an indicator of the distance that person to the sensor and try improve accuracy  by considering a minimal size of the people detected. Thus, bounding boxes smaller than a parameter threshold are ignored. \added{
Another difficulty regards the differentiation among real people, cyclists and disabled people. The tested methods detect people even in the case they are sitting or riding.
}

We utilized the deep-learning based method R-FCN~\cite{dai16rfcn} over the entire dataset to obtain a database with the number of person detected in each image. This database is then aggregated in space and time to create a visualization of the person counts by finding the average number of person per image in each region. 


\subsection{Sensing model}
\label{sec:person_and_sensors_flow_model}

\begin{figure}
    \centering
    \shadowimage[width=0.55\textwidth]{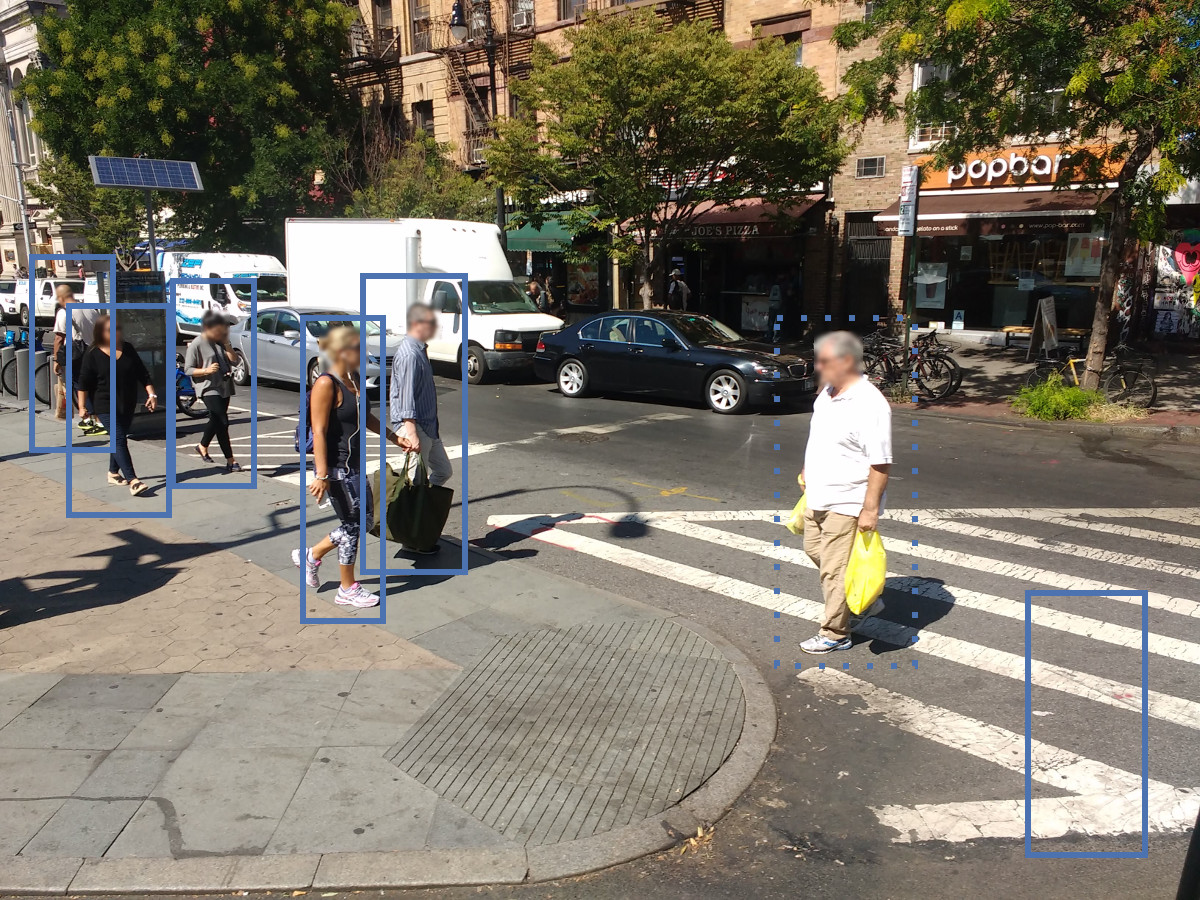}
	\caption{Hypothetical person detection. The person on the center was not identified by the detector and is a \emph{false-negative}. The rightmost detection is a \emph{false-positive}. Underlying image provided by Carmera.}
    \label{fig:detection}
\end{figure}

As current person detection algorithms are far from perfect, it is natural to wonder about the accuracy of any person count resulting from their use. In this section we provide a theoretical analysis of the effect of algorithmic errors on the final count.    

In our model, we assume that the world is modeled by a number of small regions, or buckets, each of which we intend to measure a density. Sensors and people move around a world in some random fashion. At regular intervals, each sensor takes an independent measurement of the nearby person count and updates the recorded density at its current location, $x$. More formally, each time a sensor takes a sample, it obtains a measurement represented by the random variable $N\left(x\right)$. While we do not specify the distribution of  $N\left(x\right)$, we assume that the expected value follows the formula

\begin{equation}
E\left[N_i\left(x\right)\right] = p n_i(x) + \lambda  
\label{eq:sample_process}
\end{equation}

Here $n_i\left(x\right)$ is the actual number of people in the location and time being sensed. $p$ is a number giving the success rate of the vision algorithm and $\lambda$ indicating its false positive rate.

The result of this process is the density of people at each location, $\psi\left(x\right)$.

\begin{equation}
\psi\left(x\right) = \frac{1}{k}\sum_i N_i\left(x\right)
\label{eq:sampling_values}
\end{equation}

For comparison, the ground truth density $\phi\left(x\right)$, defined respectively by (where $k$ is the number of steps and  samples),

\begin{equation}
\phi\left(x\right) = \frac{1}{k}\sum_i{n_i(x)}
\label{eq:true_values}
\end{equation}

We show in Appendix \ref{sec:theoretical-asymptotic-error}, Equation \ref{eq:expected_value_calulation} that the expected value of $\psi\left(x\right)$ is

\begin{equation}
E\left[\psi\left(x\right)\right] =  p\phi\left(x\right) + \lambda 
\label{eq:expected_phi_values_in_paper}
\end{equation}

In other words, $\psi\left(x\right)$ is a biased estimator of $\phi\left(x\right)$. Unless our sensing algorithm precisely follows Equation \ref{eq:expected_phi_values_in_paper}, we are unable to transform this biased estimator into an unbiased one. Furthermore, even in the ideal case, $p$ and $\lambda$ may not be known. Instead, we directly utilize $\psi\left(x\right)$ and attempt to find a relative histogram. That is, we expect to get a number proportional to the density of the number of people at a location and not the actual density.  As such, for any constant $a$, our density is equivalent to one scaled to $\psi'\left(x\right)=a\psi\left(x\right)$. Treating the distribution as a vector, we measure the direction but not the magnitude.  In the terminology of group theory, our measurement suggests a density within the equivalent class:

\begin{equation}
\left[\psi\right] = \left\{ a \in \mathbb{R}_+  \mid a\psi \right\}
\end{equation}

To validate our measurement we need a metric that indicates how well the equivalent class compares to the ground truth distribution $\phi\left(x\right)$. To do that, we compare the ground truth to the unique closest element within the equivalent class. As a vector projection, this minimum element is (see Appendix \ref{sec:proof-of-metric-formula} for a proof):

\begin{equation}
\psi' =
  \begin{cases}
    \psi \frac{<\psi,\phi>}{|\psi|^2} & \quad |\psi| \neq 0\\\
    0      & \quad |\psi| = 0
  \end{cases}
\label{eq:psi_prime_def}
\end{equation}

\noindent which we can then compare using the usual euclidean metric $|\psi'-\phi|$. However, this metric depends on the number of locations in the map, as well as the number of people. As such, we normalize the metric to between 0 and 1, to obtain a final metric:

\begin{equation}
    \frac{|\psi'-\phi|}{|\psi'|+|\phi|}
    \label{eq:error_metric}
\end{equation}




In Appendix B, we find a bound additionally parameterized by $h > 0$ (the average density of people) and $c \ge 1$ (a function of the distribution of $\phi$).  Both of these parameters depend on the resolution of the heat map in addition to person distribution. In many cases $c$ can not be determined, as such we can use the inequality in Equation \ref{eq:metic_ineq} of Appendix \ref{sec:theoretical-asymptotic-error} to conclude that over long periods of time:

\begin{equation}
\lim_{k\to\infty} \frac{|\psi'-\phi|}{|\psi'|+|\phi|}  \leq \frac{\sqrt{{{c}^{2}}-1}\lambda}{2{{c}^{2}}hp} \leq \frac{1}{4}\frac{\lambda}{hp} 
\label{eq:full_inequality}
\end{equation}

It is important to note that $h$ represents the ground truth density of people, in the same units of $\phi$. If only the sampled average density, $\hat{h}$, is know, the unbiased estimator of $h$, $\frac{\hat{h}-\lambda}{p}$ can be used. This leads to the bounds

\begin{equation}
\frac{1}{4}\frac{\lambda}{hp}   \approx  \frac{1}{4}\frac{\lambda}{\bar{h}-\lambda} 
\label{eq:full_inequality_h_bar}
\end{equation}

This final formula is only dependent on the false positive rate of the sensing algorithm and the average density of sensed objects measured by process, making it suitable for practical sensing applications.  We wish to emphasize that this inequality is true whenever Equation \ref{eq:expected_phi_values_in_paper} holds regardless of the underling probability distribution. This function is only useful when $\lambda \le \bar{h}$ and in that domain, it is a monotonically increasing function of $\lambda$.

\subsection{Simulation}
\begin{figure}
    \centering
    \begin{subfigure}[b]{0.45\textwidth}
        \includegraphics[width=\textwidth,valign=c]{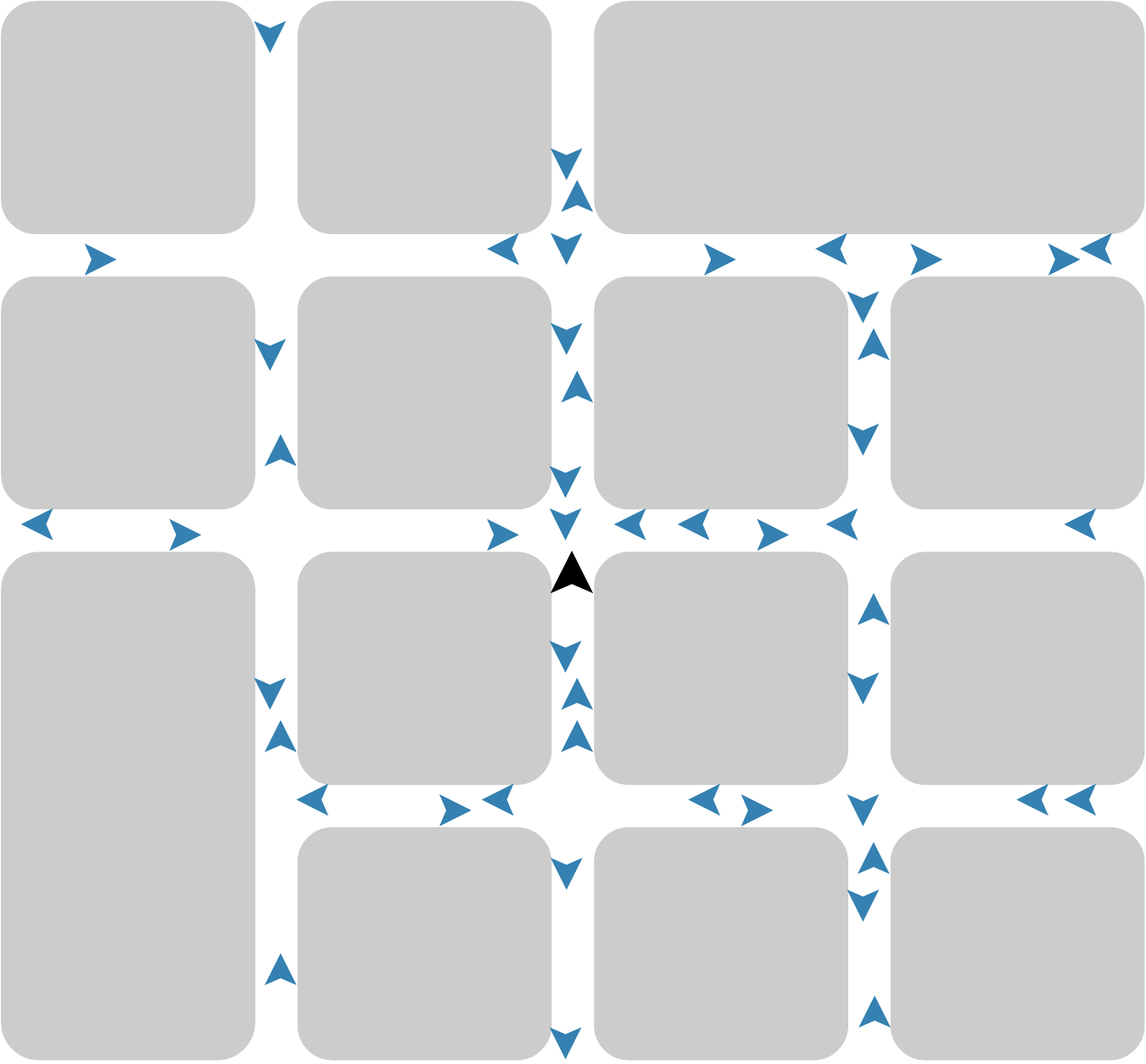}
    \end{subfigure}
	\qquad
    \begin{subfigure}[b]{0.45\textwidth}
         \centering
        \includegraphics[width=.75\textwidth,valign=c]{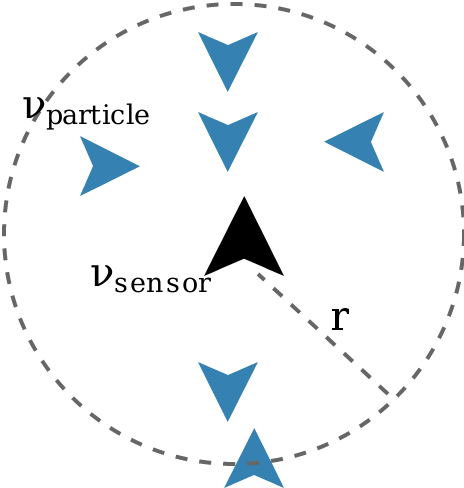}
    \end{subfigure}
        \caption{Left: An illustration of our simulation containing a sensor (center) moving through an environment with numerous person. Right: Each sensor moves with uniform speed $\nu_{s}$ and is able to sense people within a radius of $r$. Each sensing operation has a probability $p$ of correctly detecting each person and, on expectation, finds $\lambda$ false positives.  People move with uniform speed $\nu_{p}$. }
        \label{fig:sensor}
\end{figure}

The real-life acquisition process lacks some of the simplifications we used in our model. For example, samples taken in spatial and temporal proximity are correlated. To examine the performance of the sensing systems in the face of these non-ideal circumstances, we created a \emph{discrete event simulation}~\cite{law2007simulation} to compare sensed distributions to a known ground truths.

As illustrated in Figure~\ref{fig:sensor}, we simulated a number of mobile sensors that detect nearby particles. \added{Sensors simulate the person detection event and their movement simulate the movement of the camera-equipped fleets responsible for the acquisition of data}. Each sensor has a circular coverage of radius $r$. Collision among particles and sensors are ignored for simplicity. Sensors and particles move with uniform speeds $\nu_{sensor}$ and  $\nu_{particle}$ respectively. The simulation world is mapped as a graph, as in Tian et al.~\cite{tian2002graph}. Each node in the graph is a traversable point by both sensors and particles and edges represent a path between the end nodes.

We assume that, in each time step, sensor has an independent chance, $p$, of detecting each of the $n\left(x\right)$ person within range along with an independent chance per location to obtain a false positive. These assumptions lead to $N\left(x\right)$ being sampled from the sum of a binomial distribution with mean $p$ and a Poisson process with a given expected number $\lambda$.  A calculation of the expected value indicates that Equation \ref{eq:sample_process} is satisfied and within the proposed bounds.

 \begin{algorithm}
 Initialization(map, currentposition, destination)\;
 \While{indefinitely}{
  \eIf{destination = $\emptyset$}{
   destination $\leftarrow$ random(map)\\
   path $\leftarrow$ A*(currentposition, destination, map)
   }{
   currentposition $\leftarrow$ pop(path)\;
  }
 }
 \caption{Mobility model of sensors and particles of the simulation.}
 \label{alg:mobility}
\end{algorithm}

The system state can be described by various state variables: sensors and particles positions, sensors and particles waypoints, real density of particles and sensed density of particles.  Sensors and particles move with a variation of the random waypoint model~\cite{johnson1996dynamic}, differing to it by the fact that sensors and particles are not allowed to change speeds; they have fixed speed given by the system parameters $\nu_{sensor}$ and $\nu_{particle}$. When a new destination is randomly picked, the trajectory on the map graph is computed using the A* algorithm~\cite{hart1968formal} and the points of the trajectory are pushed to a heap (please refer to Algorithm~\ref{alg:mobility}). 

As time progresses we obtain a 2D histogram for the sensed density as well as the ground truth density of particles. We are primarily interested in the difference between them, as given by the metric in Equation \ref{eq:error_metric}.

%% file: experiments.tex
\section{Experimental results}
\label{sec:experiments}
\added{This section is divided into three parts: person detection and density map generation; calibration of the simulation; and simulation. The full source code is public available}~\footnote{\url{https://github.com/VIDA-NYU/pedestrian-sensing-model}}.
\subsection{Person detection results}
\label{subsec:detection}

We manually tagged $600$ images to use as a ground truth. 
The R-FCN algorithm~\cite{dai16rfcn} was trained on the Pascal VOC 2007 dataset with the 101-layers neural network architecture proposed by He et al.~\cite{he2016deep}. We evaluated the method with detection scores ranging from $0.0$ to $1.0$, spaced by $0.1$. 

\begin{figure}
    \centering
    \begin{subfigure}{0.45\textwidth}
        \centering
        \includegraphics[width=\textwidth]{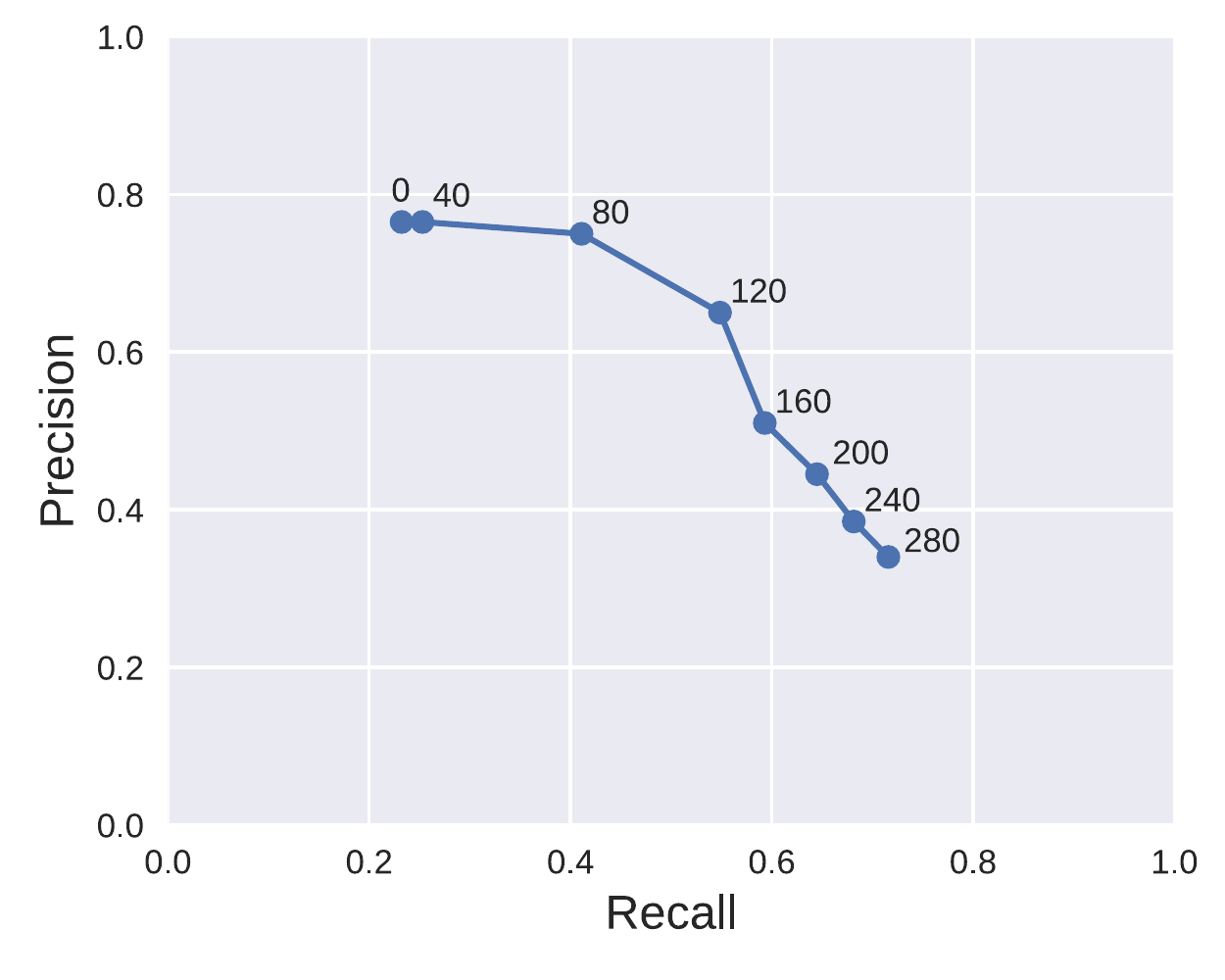}
        \end{subfigure}
        \qquad
    \begin{subfigure}{0.45\textwidth}
        \centering
        \includegraphics[width=\textwidth]{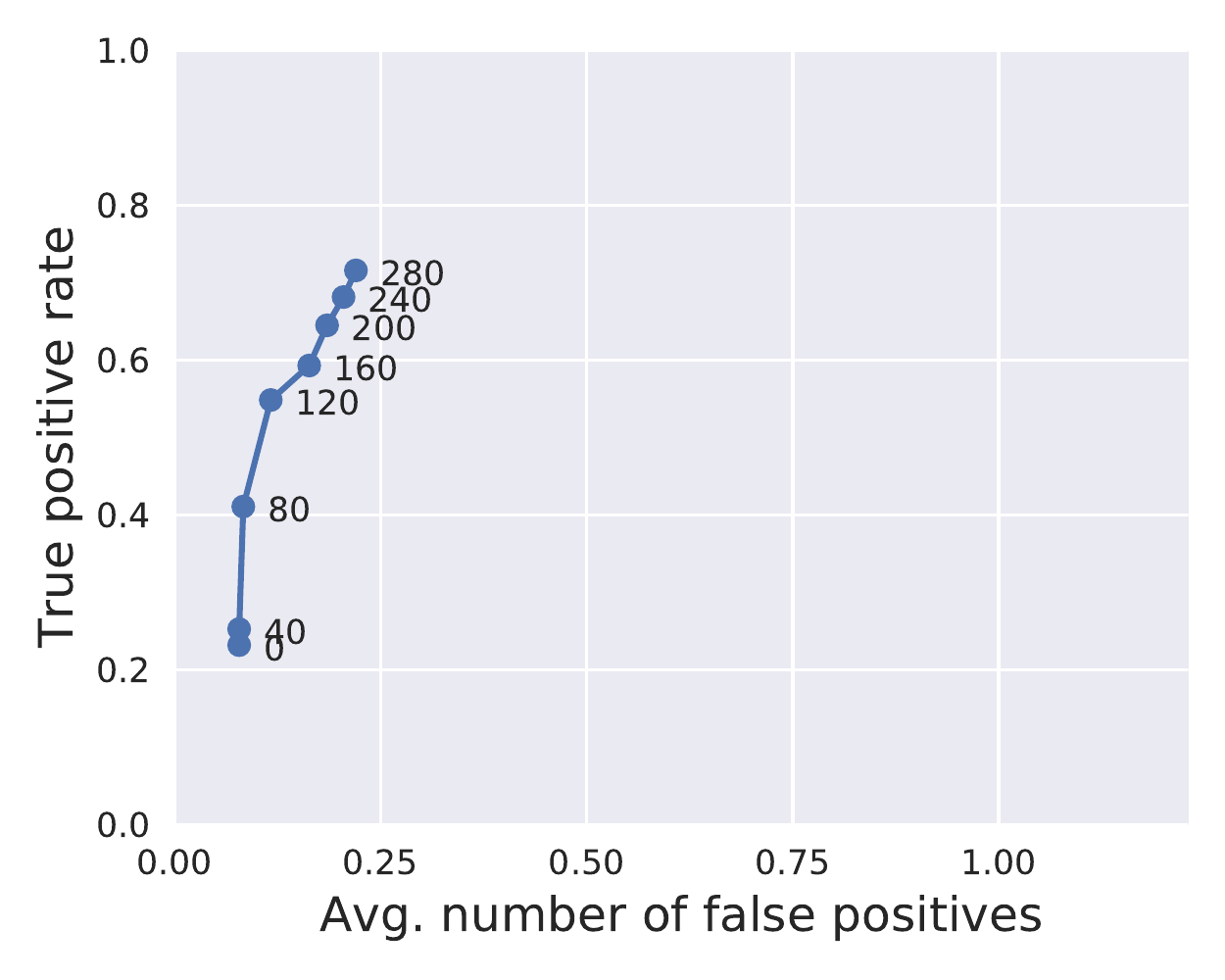}
        \end{subfigure}
        \caption{Evaluation of R-FCN~\cite{dai16rfcn} for different ground-truth height thresholds. The utilized model has a Resnet-101 backbone ~\cite{he2016deep} trained on the Pascal VOC 2007 dataset~\cite{everingham2010pascal}.}
        \label{fig:graphminheight}
\end{figure}

	None of evaluations achieved recalls exceeding $80\%$ (see Figure~\ref{fig:graphminheight}) and this fact is inherent to the difficulty of object detectors in detecting small objects as discussed in Section~\ref{subsec:cv}. To mitigate such issue, the detection model we propose assumes a finite radius of coverage (see Figure~\ref{fig:sensor}) and thus, we establish a limit on the size of the objects detected in the image. Figure~\ref{fig:graphminheight} shows the results of the adopted detector over our sample as we vary the minimum acceptable height. As we can see, the higher the ground-truth height threshold, the higher the precision and specially the recall of the method.


\begin{figure}
    \centering
        \includegraphics[width=.9\textwidth]{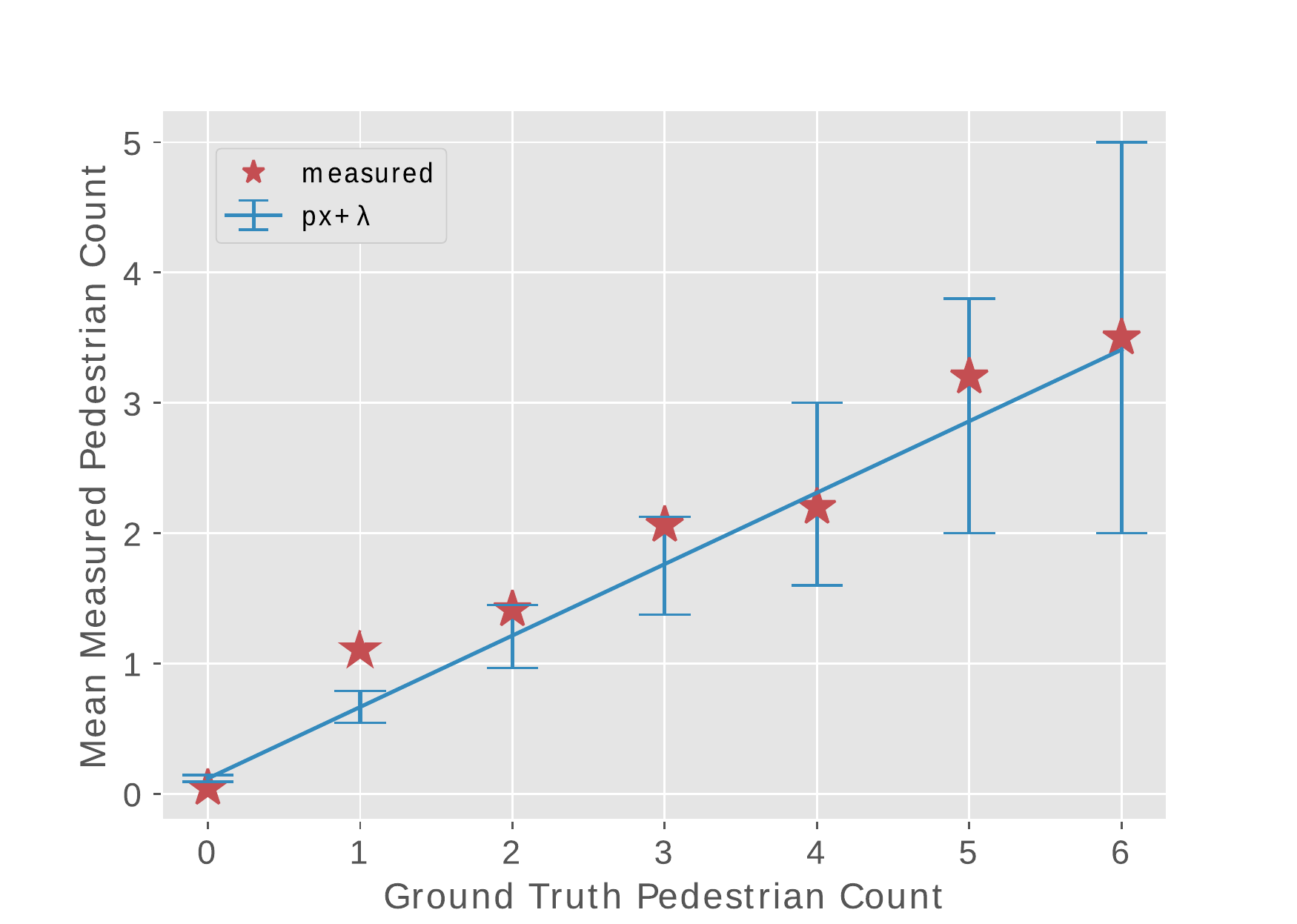}
        \caption{A comparison of the ground truth and the measured person count from the $600$ tagged test images. While the actual true positive and false positive counts do not match the expected statistics (left), the total measured person count can be close to approximated as linear (right). It should be noted that this is only an approximation as, even taking sampling errors into account, the mean measured count do not fit a linear model. Error bars are the $95\%$ confidence interval of the mean, calculated by assuming the sampling process described in Section \ref{sec:person_and_sensors_flow_model}. }
	\label{fig:linearfit}
\end{figure}

We utilized as our detector the R-FCN method~\cite{dai16rfcn} with a residual network of 101 layers architecture~\cite{he2016deep} trained on Pascal VOC 2007~\cite{everingham2010pascal}. The model was trained using a weight decay of $0.0005$ and a momentum of $0.9$. Assuming a method minimum score of $0.7$ and height threshold of 120 pixels, overall $7,474,623$ people were detected.  

\begin{figure}
    \centering
	\includegraphics[width=0.8\textwidth]{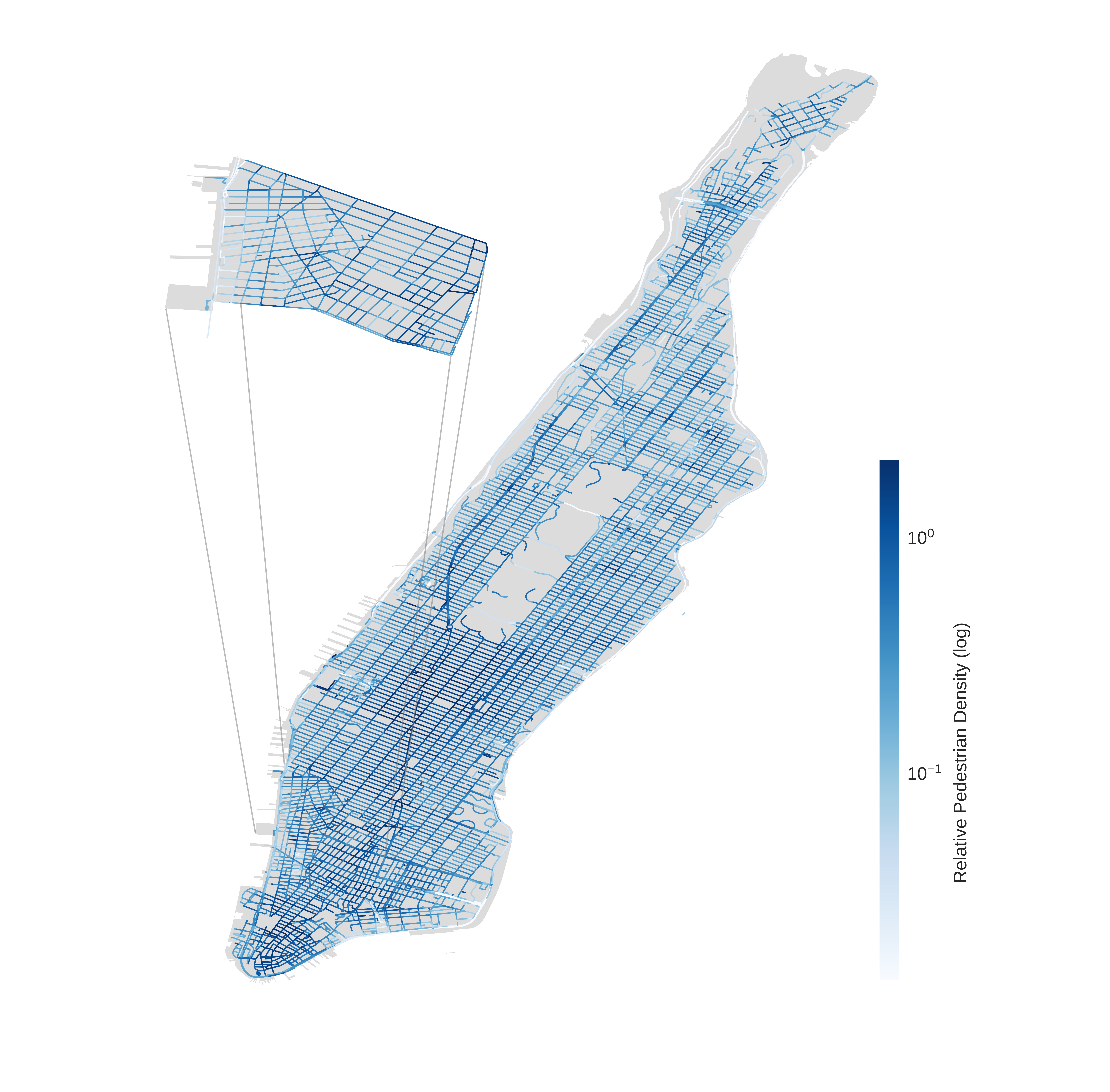}
	 \caption{\added{Visualization of the density of people in Manhattan. The scale of colors represent the relative density of people. The main figure depicts the heatmap over the island of Manhattan. The same heatmap is enlarged in the top-left of the figure to show the details of midtown and surrounding areas. \textit{Underlying map data taken from OpenStreetMap \cite{openstreetmap}. Not drawn to scale.}}\label{fig:visionmanhattanheatmap} }
\end{figure}

As shown in Figure \ref{fig:linearfit}, the total number of people detected is approximately linear satisfying the requirements of the model proposed in Section \ref{sec:person_and_sensors_flow_model}, despite stastically significant deviations. These stem from the visions algorithm's better than expected performance for images without any person and worse than expected performance for images with a single person. While we do not know how these deviations would effect the error bounds given in Equations \ref{eq:full_inequality} and \ref{eq:full_inequality_h_bar}, we hypothesize that the two deviations would cancel themselves out and bound may still approximately hold with a slightly larger equivalent $\lambda$.

A visualization of the density of people in entire Manhattan can be seen in Figure~\ref{fig:visionmanhattanheatmap}. \added{For each analyzed picture, the number of people is determined using the detection algorithm and these counts are aggregated by streets. The topology of the streets were obtained from OpenStreetMaps~\cite{openstreetmap}. The samples were organized in a KD-tree~\cite{bentley1975multidimensional} and aggregated by street segments. To account for the imbalance of the number of images per segment, the average of the counts is considered.} 

Distribution of the number of people, like ours, can be useful for city planing, commercial, and other purposes. Taxis seeking riders, food trucks seeking customers, and businesses seeking storefronts all benefit from large crowds. However, traffic and self-driving cars do not. A knowledge of the  distributions of the number of people can allow city planers, civil engineers, and traffic engineers to make better decisions. 

Our person map can also show the effect that features of the city have on it's people. In addition to populated neighborhoods, subway stations, and attractions like the Metropolitan Museum of Art are all associated with a spike in the person densities. These spikes might be too localized to be detected using traditional methods. Further studies of vision based person counts may lead to a better understanding of the interplay between a city's environment and it's occupant's walking habits. 

\subsubsection*{\added{Special cases}}
\begin{figure}[ht]
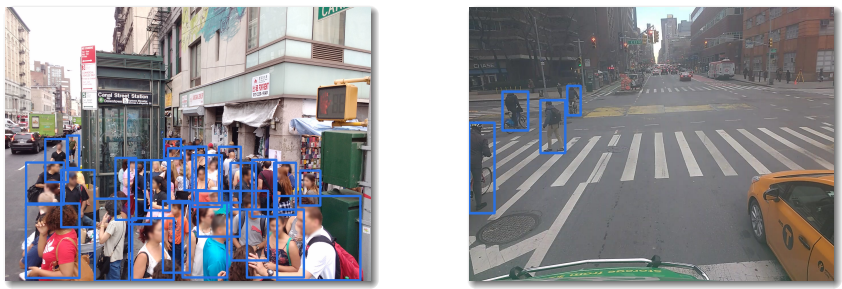

		\centering
		\shadowimage[width=0.4\textwidth]{crowd/1284182.jpg}\qquad
		\shadowimage[width=0.4\textwidth]{{dr5rsxfkfsw6-dr5rsxfktk1g-cds-44290495868b6534-20170201-0907.mp4-8173_bicycles}.jpg} 
		\caption{\added{Person detection by the method in special cases. On the left, person detection in a crowded scene. On the right, pedestrians and non-pedestrians (a cyclist) are equally identified by the method. }}
		\label{fig:bicycles}
	\end{figure}

		\added{A visual inspection of the detections showed that the methods are scale-sensitive, having difficulty to detect tiny-scale objects. However, when the objects have representative sizes, even in crowded scenes, the method proved to be robust. Another difficulty of the method regards the differentiation between pedestrians and non-pedestrians, such as cyclists and disabled people (see Figure~\ref{fig:bicycles}). Such differentiation would require a novel computer vision method that considered the context of the scene. We performed a manual inspection over our annotated dataset and cyclists represented $2.4\%$ of all annotated pedestrians. This low rate supports the argument that such limitation of the method does not invalidate the results statistically.}

\begin{figure}
    \centering
    \begin{subfigure}[b]{0.9\textwidth}
        \includegraphics[width=\textwidth]{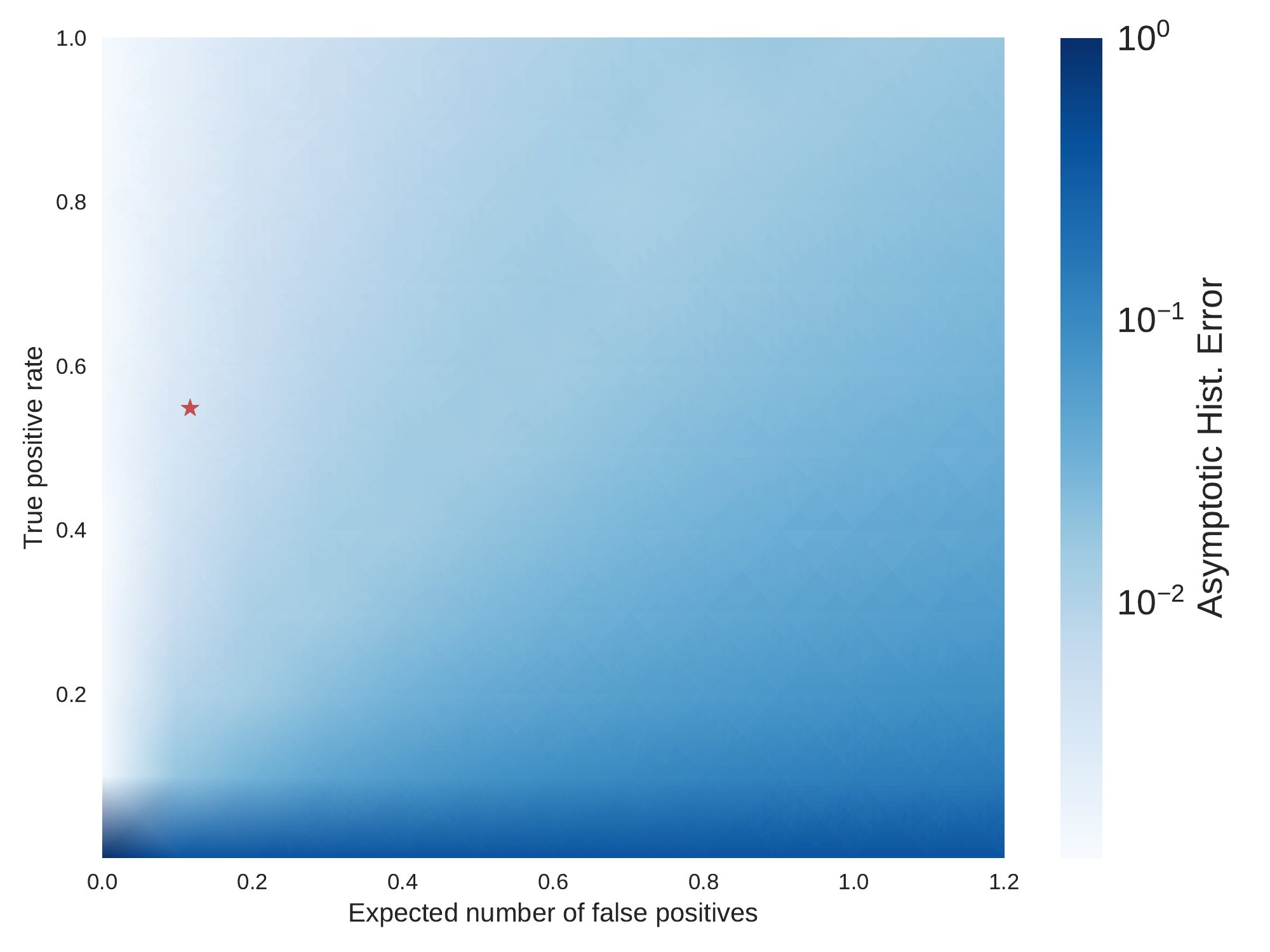}
    \end{subfigure}
        \caption{Asymptotic error metric between the sensed and ground truth histograms in our simulation, plotted as a function of the sensors true positive and expected number of false positives. The star in red depicts the results of the vision algorithm utilized with an asymptotic error of $0.062$.}
        \label{fig:calibration}
\end{figure}

\added{
\subsection{Simulation calibration}
	Simulation parameters have been calibrated according to the computer vision experiments. In Figure~\ref{fig:calibration} we can see in red the expected number of false positives and the true positive rate of the computer vision method utilized.  Given the average people density ($\hat{h}$) of 0.587 obtained in the previous section, following Equation~\ref{eq:full_inequality}, an error of $0.062$ is obtained.  
	}

\added{
We compared the number of measured person counts as a function of the average number of ground truth people in each of the 600 manually labeled images to test the linear assumption used in Equation \ref{eq:expected_phi_values_in_paper}. Error bars for the mean were computed using the $5\%$ to $95\%$ values of the median of the appropriate sample process given in section \ref{sec:person_and_sensors_flow_model}. We measured the true positive rate ($p$) to be $0.54$ and the average number of false positives ($\lambda$) to be of $0.117$. 
}

\FloatBarrier
\subsection{Simulation results}

\begin{table}
	\small
        \caption{ Parameters of the simulation along with the values we used in the experiments.  }
        \centering

        \begin{tabular}{lrr}
        \toprule
		\textbf{Parameter} & \textbf{Symbol} & \textbf{Values}\\ \midrule
Number of people &$N_{people}$ & 50000 \\
Person speed &$\nu_{person}$ & 1 \\
Sensor speed &$\nu_{sensor}$ & 3\\
Number of sensors &$N_{sensors}$ & 10000\\
		Sensor true positive rate&$p$  & \{0.0, 0.1, 0.2, 0.3, 0.4, 0.5,\\
		&  & 0.6, 0.7, 0.8, 0.9, 1.0\}\\
		Sensor exp. number  of false positives&$\lambda$ & \{0.0, 0.1, 0.2, 0.3, 0.4, 0.5, 0.6,\\
		& & 0.7, 0.8, 0.9, 1.0, 1.1, 1.2\}\\
Sensor range &$r$ & 1\\
        \bottomrule
        \label{tab:parameters}
        \end{tabular}
\end{table} 

We evaluated different true positive rates and expected number of false positives of the sensors, $p$ and $\lambda$ and we used a Mersenne Twister pseudo number generator~\cite{matsumoto1998mersenne}. The various values for the parameters used in our experiments are listed in Table~\ref{tab:parameters}. For the 143 possibilities combination of values, we ran the simulation for $20,000$ time steps $20$ independent times.  

The code is primarily implemented in Python with performance-sensitive sections implemented in Cython~\cite{behnel2010cython}.  The average time to run a single experiment of this optimized code is of $11,718$ seconds. Running the experiments on a single machine would take roughly one year of computation. By running them in parallel, it took 11 days.

For each experiment we examine the decay of the metric given by Equation \ref{eq:error_metric} as a function of the cumulative number of samples captured by all the sensors. We assume the error continues to decay until it reaches an asymptotic minimum error within the $20,000$ simulation time steps. Afterwards, we take the average decay curve of all $20$ runs for each settings configuration and take the average of the last $200$ values to find the asymptotic value.

\begin{figure}
    \centering
    \begin{subfigure}[b]{0.9\textwidth}
        \includegraphics[width=\textwidth]{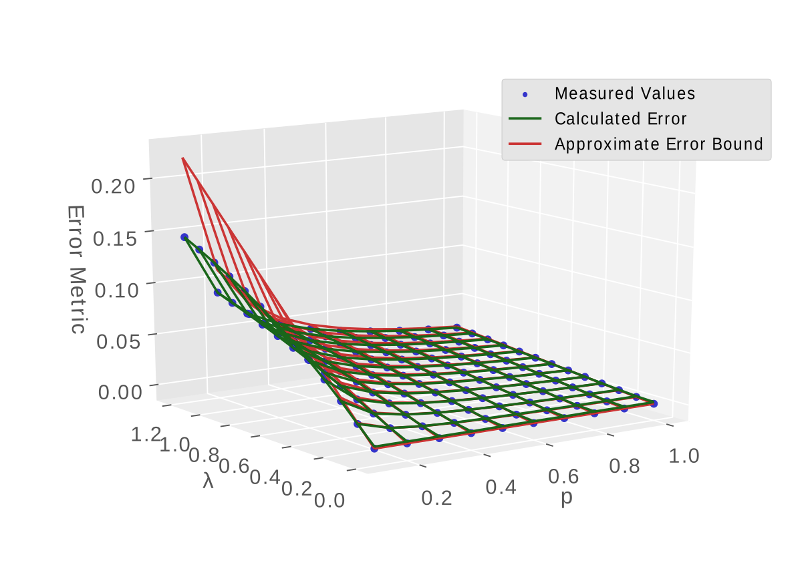}
    \end{subfigure}
	\caption{Comparison of the asymptotic histograms error measured from the simulation experiments (blue dots)  to the theoretical close form function and the approximate bound (please refer to Equation \ref{eq:full_inequality}), the red mesh. }
        \label{fig:errorsheatmap}
\end{figure}


We can visualize the results from the simulations in  Figure~\ref{fig:errorsheatmap} which shows how the variation of true positive rate (p) and number of false positives ($\lambda$) affect our histogram error. If we take a profile of say $0.2$ of true positive rate ($p=0.2$) we can see how the errors are greatly affected by the variation of the expected number of false positives, varying from very low to high error values (represented by the variation on the color saturation). We compare these values to our theoretical formulas (see Equation~\ref{eq:full_inequality}), and show they are approximately equal. Finally, We show that they are within the bound given by Equation~\ref{eq:full_inequality} (red mesh).

\FloatBarrier

%% file: conclusion.tex
\section{Conclusion}

In this project we applied computer vision techniques over a large and novel set of street-level images to obtain a pedestrian density map of the region of Manhattan. Inspired by the density estimation procedure we proposed a sensing model and established theoretical bounds for the density estimation error. Results give evidence that even when considering faulty sensors such as abstracted by our detection algorithms, reliable density maps can be obtained under our model assumptions.

Besides the results presented, there are other potential future avenues of studies as discussed next. First, we should caution that any application of our methodology should perform statistical tests to ensure that their results are statistically significant. While we set bounds on the asymptotic error after the sampling process converges, we have only provided case studies and heuristics for the time to convergence. It would be interesting to find a formal bound on time to convergence as well as provide guidelines for the appropriate statistical tests to validate the data post collection.

\added{Our experiments could be extended to consider alternative mobility models~\cite{camp2002survey}, higher performance computer vision approaches, possibly semi-supervised ones~\cite{chapelle2009semi,tokuda2018a} as well as dynamics models including macroscopic ones~\cite{helbing1998fluid, iwata2017estimating}.}  We could also use data completion algorithms~\cite{ gandy2011tensor, li2017city} or semi-supervised approaches~\cite{zhu2005semi} to reconstruct a city-wide people map.

The people distribution generated will then be able to be combined  with other urban datasets such from Socrata~\cite{nycopendata}, weather, crime rate, vandalism~\cite{tokuda2019quantifying}, census data, public transportation~\cite{silva2019integrated}, bicycles and shadows~\cite{miranda2018shadow}. We additionally aim to explore apparently disparate datasets such as from wind and from garbage collection.

Another future work is incorporation of advances such as from Microsoft Inc.~\cite{photosynth} to visualize our images in the context of the city and use this visualization to gain additional insights into other datasets. 

Additionally, we hope to use 3D pop-ups and/or photo based rendering to fully enhance the images in the three dimensional environments.

%% file: ack.tex
\section*{Acknowledgements}

We thank Carmera for their collaboration. We also thank Harish Doraiswamy, Fabio Miranda, Alexandru Telea and Helio Pedrini for providing insights, comments, and suggestions that greatly contributed to this work.
This work was supported in part by: NSF awards CNS-1229185, CCF-1533564, CNS-1544753, CNS-1730396, CNS-1828576; FAPESP (grants \#14/24918-0 and \#2015/22308-2); CNPq and CAPES; the Moore-Sloan Data Science Environment at NYU, and C2SMART. C.~T.~Silva is partially supported by the DARPA D3M program. Any opinions, findings, and conclusions or recommendations expressed in this material are those of the authors and do not necessarily reflect the views of DARPA.

%% file: appendix.tex
\appendix
\section{Proof of metric formula}
\label{sec:proof-of-metric-formula}

Here we derive the formula for the closest point in our class and the ground truth vector. This is equivalent to solving:

\begin{equation}
\min_{a} |a \psi - \phi|^2
\end{equation}

First we expand the distance metric using the euclidean inner-product:

\begin{align}
|a \psi - \phi|^2  &= <a \psi - \phi, a\psi - \phi> \\
                    &= a^2<\psi, \psi> - 2a<\psi, \phi> + <\phi,\phi>
\end{align}

which is minimized by 

\begin{align*}
   a = \frac{<\psi, \phi>}{<\psi, \psi>} = \frac{<\psi, \phi>}{|\psi|^2}
\end{align*}

when $|\psi|^2 \neq 0$. For $|\psi|^2 = 0$, then $\psi=0$ and

\begin{equation}
\min_{a} |a \psi - \phi|^2 = |\phi|^2
\end{equation}
regardless the value of a.

Substituting this value into $a$, we obtain our equation above:

\[ \psi' =
  \begin{cases}
    \psi \frac{<\psi,\phi>}{|\psi|^2}      & \quad |\phi| \neq 0\\\
    0      & \quad |\phi| = 0
  \end{cases}
\]

When $<\psi, \phi> = 0$ then $\psi'=0$ and we get the same value. 

\section{Theoretical Asymptotic Error}
\label{sec:theoretical-asymptotic-error}

In this paper we will derive the equation for sensor error that we give in Equation \ref{eq:full_inequality}. The error bounds only assume that $E\left[\psi\left(x\right)\right]=p\phi\left(x\right) + \lambda $ for some values of $p$ and $\lambda$. First, we will show that, for the simulation, the values of $p$ and $\lambda$ agree with the parameters of the same name. 

As noted in Equation \ref{eq:sampling_values}, the sampling process for the simulation results in the following sampled values for each location:  
\[
\psi\left(x\right) = \frac{1}{k}\sum\left( T(n(x)) + F\right)
\]

\noindent where $T(n(x))$ is a sampled from a binomial distribution with mean $n(x)$ and $F$ is a Poisson process~\added{\cite{kleinrock1976queueing}} with a mean of $\lambda$.

At the same time, the ground truth distribution of people at each location is given by Equation~\ref{eq:true_values}  
\[
\phi\left(x\right) = \frac{1}{s}\sum{n(x)}
\]
In this appendix we will make the simplifying assumption that $s=k$

The executed value of the sampled can then be found by (noting that the random variables are all independent)

\begin{equation}
\begin{split}
E\left[\psi\left(x\right)\right] & = \frac{1}{s}\sum\left( E\left[T(n(x))\right] + E\left[F\right]  \right) \\
& = \frac{1}{s}\sum\left( p n(x) + \lambda  \right) \\
& = p\frac{1}{s}\sum\left( n(x)\right)  + \lambda \\
& = p\phi\left(x\right) + \lambda 
\end{split}
\label{eq:expected_value_calulation}
\end{equation}

From this point on, all results will only depend on the equations $E\left[\psi\left(x\right)\right]=p\phi\left(x\right) + \lambda $ and not the underlying sampling process. 

Let $m$ be the total number of people and $r$ be the sampling location. In a real world scenario, $m$ may not be well defined. As such, we will work in terms of $h=\frac{m}{r}$, the density of people. 

We want to calculate $\frac{|\psi'-\phi|}{|\psi'|+|\phi|}$ and we are going to compute each term separately.

By the law of large numbers, in the limit of $s\to\infty$, $\psi$ approaches 

\[
\psi\left(x\right) \to E\left[\psi\left(x\right)\right]  =  
\phi\left(x\right)p + \lambda
\]


Using this limit, we can find the asymptotic value of $\psi'$, as defined by Equation \ref{eq:psi_prime_def}, can be found:

\begin{equation}
\begin{split}
\psi' & = \psi \frac{<\psi,\phi>}{|\psi|^2} \\
& \to \left(p\phi + \lambda \right) \frac{\left<p\phi + \lambda,\phi\right>}{|p\phi + \lambda|^2}  \\
& = \left(p\phi + \lambda \right) \frac{p\left|\phi\right|^2 + \lambda\left<1 ,\phi\right>}{ p^2\left|\phi\right|^2 + 2p\lambda\left<1 ,\phi\right> + \lambda^2\left<1,1\right> } \\
& = \left(p\phi + \lambda \right) \frac{p\left|\phi\right|^2 + m\lambda}{ p^2\left|\phi\right|^2 + 2mp\lambda + r\lambda^2 } \\
& = \left(\phi + \frac{\lambda}{p} \right) \frac{p^2\left|\phi\right|^2 + mp\lambda}{ p^2\left|\phi\right|^2 + 2mp\lambda + r\lambda^2 }
\end{split}
\label{eq:psi_prime_calulate_limit}
\end{equation}

Here, $\left<\cdot,\cdot\right>$ is the Euclidean inner product and $1$ is the vector with all ones. Note that $\left<1 ,\phi\right> = m$ and $\left<1,1\right>=r$

The magnitude of $\psi'$ can then be found by
\begin{equation}
\begin{split}
\left| \psi'\right| & =  \frac{p^2\left|\phi\right|^2 + mp\lambda}{ p^2\left|\phi\right|^2 + 2mp\lambda + r\lambda^2 }\sqrt{ \left<\phi,\frac{\lambda}{p}1\right> } \\
& = \frac{p^2\left|\phi\right|^2 + mp\lambda}{ p^2\left|\phi\right|^2 + 2mp\lambda + r\lambda^2 }\sqrt{\left|\phi\right|^2 +\frac{r\lambda^2}{p^2} + 2\frac{m\lambda}{p}  } \\
&  = \frac{p\left|\phi\right|^2 + m\lambda}{ p^2\left|\phi\right|^2 + 2mp\lambda + r\lambda^2 }\sqrt{p^2\left|\phi\right|^2 +r\lambda^2 + 2mp\lambda }
\end{split}
\label{eq:magnitude_of_psi_prime}
\end{equation}

Similarly, the difference between $\psi'$ and the ground truth sampling can be found by

\begin{equation}
\begin{split}
\psi'-\phi & = \left(\phi + \frac{\lambda}{p} \right) \frac{p^2\left|\phi\right|^2 + mp\lambda}{ p^2\left|\phi\right|^2 + 2mp\lambda + r\lambda^2 } - \phi \\
& = \left(\phi + \frac{\lambda}{p} \right) \frac{p^2\left|\phi\right|^2 + mp\lambda}{ p^2\left|\phi\right|^2 + 2mp\lambda + r\lambda^2 } - \phi\frac{p^2\left|\phi\right|^2 + 2mp\lambda + r\lambda^2}{p^2\left|\phi\right|^2 + 2mp\lambda + r\lambda^2} \\
& = \phi \frac{p^2\left|\phi\right|^2 + mp\lambda-p^2\left|\phi\right|^2 - 2mp\lambda - r\lambda^2}{ p^2\left|\phi\right|^2 + 2mp\lambda +  r\lambda^2 } + \frac{\lambda}{p}\frac{p^2\left|\phi\right|^2 + mp\lambda}{ p^2\left|\phi\right|^2 + 2mp\lambda + r\lambda^2 } \\
& = -\phi \frac{  mp\lambda + r\lambda^2}{ p^2\left|\phi\right|^2 + 2mp\lambda +  r\lambda^2 } + \frac{\lambda}{p}\frac{p^2\left|\phi\right|^2 + mp\lambda}{ p^2\left|\phi\right|^2 + 2mp\lambda + r\lambda^2 } \\
& =  \frac{\lambda}{ p^2\left|\phi\right|^2 + 2mp\lambda +  r\lambda^2 } \left( -\phi\left( mp + r\lambda\right) + p\left|\phi\right|^2 + m\lambda  \right)
\end{split}
\label{eq:psi_prime_minis_phi_calulate_limit}
\end{equation}

The magnitude of which can be found by
\begin{equation}
\begin{split}
\left| \psi'-\phi\right| &  = \frac{\lambda}{ p^2\left|\phi\right|^2 + 2mp\lambda +  r\lambda^2 }\sqrt{\left< -\phi\left( mp + r\lambda\right), \left(p\left|\phi\right|^2 + m\lambda\right)1 \right>} \\
&  = \frac{\lambda}{ p^2\left|\phi\right|^2 + 2mp\lambda +  r\lambda^2 } \sqrt{\left|\phi\right|^2\left(mp + r\lambda\right)^2 + \left(p\left|\phi\right|^2 + m\lambda  \right)^2r - 2\left(mp + r\lambda\right)\left(p\left|\phi\right|^2 + m\lambda   \right)m  }
\end{split}
\label{eq:maginitude_psi_prime_minis_phi_calulate_limit}
\end{equation}

Equations \ref{eq:magnitude_of_psi_prime} and \ref{eq:maginitude_psi_prime_minis_phi_calulate_limit} can be used to find our metric as defined by Equation \ref{eq:error_metric} 
\begin{equation}
\begin{split}
\frac{|\psi'-\phi|}{|\psi'|+|\phi|} = & \frac{\frac{\lambda}{ p^2\left|\phi\right|^2 + 2mp\lambda +  r\lambda^2 } \sqrt{\left|\phi\right|^2\left(mp + r\lambda\right)^2 + \left(p\left|\phi\right|^2 + m\lambda  \right)^2r - 2\left(mp + r\lambda\right)\left(p\left|\phi\right|^2 + m\lambda   \right)m  }}{\frac{p\left|\phi\right|^2 + m\lambda}{ p^2\left|\phi\right|^2 + 2mp\lambda + r\lambda^2 }\sqrt{p^2\left|\phi\right|^2 +r\lambda^2 + 2mp\lambda } + \left|\phi\right|} \\
= & \frac{\lambda \sqrt{\left|\phi\right|^2\left(mp + r\lambda\right)^2 + \left(p\left|\phi\right|^2 + m\lambda  \right)^2r - 2\left(mp + r\lambda\right)\left(p\left|\phi\right|^2 + m\lambda   \right)m  }}{\left(p\left|\phi\right|^2 + m\lambda\right)\sqrt{p^2\left|\phi\right|^2 +r\lambda^2 + 2mp\lambda } + \left( p^2\left|\phi\right|^2 + 2mp\lambda +  r\lambda^2\right)\left|\phi\right|}
\end{split}
\label{eq:first_form_of_our_norm}
\end{equation}

However, this equation depends on $|\phi|$, $m$, and $r$ which would not be known for real applications. To account for these variables, we will introduce a new parameter, $c$:
\begin{equation}
c = \left|\phi\right|\frac{\sqrt{r}}{m}  
\end{equation}

While $c$ may also be unknown, we will be able to take a maximum of the resulting error function to get a bound. Using the bounds of L2-norm in terms of the L1-norm and noting that the L1-norm is equal to $m$, we obtain the identity

\begin{equation}
1 \leq c \leq \sqrt{r}
\end{equation}

By substituting, $\left|\phi\right| = c\frac{m}{\sqrt{r}}$ and a bit of algebra, we can transform Equation \ref{eq:first_form_of_our_norm} into a final form:

\begin{equation}
\begin{split}
\frac{|\psi'-\phi|}{|\psi'|+|\phi|}  & = \frac{\lambda \sqrt{\left|\phi\right|^2\left(mp + r\lambda\right)^2 + \left(p\left|\phi\right|^2 + m\lambda  \right)^2r - 2\left(mp + r\lambda\right)\left(p\left|\phi\right|^2 + m\lambda   \right)m  }}{\left(p\left|\phi\right|^2 + m\lambda\right)\sqrt{p^2\left|\phi\right|^2 +r\lambda^2 + 2mp\lambda } + \left( p^2\left|\phi\right|^2 + 2mp\lambda +  r\lambda^2\right)\left|\phi\right|} \\
& = \frac{\lambda \sqrt{c^2\frac{m^2}{r}\left(mp + r\lambda\right)^2 + \left(pc^2\frac{m^2}{r} + m\lambda  \right)^2r - 2\left(mp + r\lambda\right)\left(pc^2\frac{m^2}{r} + m\lambda   \right)m  }}{\left(pc^2\frac{m^2}{r} + m\lambda\right)\sqrt{p^2c^2\frac{m^2}{r}+r\lambda^2 + 2mp\lambda } + \left( p^2c^2\frac{m^2}{r} + 2mp\lambda +  r\lambda^2\right)c\frac{m}{\sqrt{r}}  } \\
& = \frac{\frac{1}{r\sqrt{r}}\lambda \sqrt{c^2\frac{m^2}{r}\left(mp + r\lambda\right)^2 + \left(pc^2\frac{m^2}{r} + m\lambda  \right)^2r - 2\left(mp + r\lambda\right)\left(pc^2\frac{m^2}{r} + m\lambda   \right)m  }}{\frac{1}{r\sqrt{r}}\left(\left(pc^2\frac{m^2}{r} + m\lambda\right)\sqrt{p^2c^2\frac{m^2}{r}+r\lambda^2 + 2mp\lambda } + \left( p^2c^2\frac{m^2}{r} + 2mp\lambda +  r\lambda^2\right)c\frac{m}{\sqrt{r}}\right)  } \\
& = \frac{\lambda \sqrt{c^2\frac{m^2}{r^2}\left(\frac{m}{r}p + \lambda\right)^2 + \left(pc^2\frac{m^2}{r^2} + \frac{m}{r}\lambda  \right)^2 - 2\left(\frac{m}{r}p + \lambda\right)\left(pc^2\frac{m^2}{r^2} + \frac{m}{r}\lambda   \right)\frac{m}{r} }}{\left(pc^2\frac{m^2}{r^2} + \frac{m}{r}\lambda\right)\sqrt{p^2c^2\frac{m^2}{r^2}+\lambda^2 + 2\frac{m}{r}p\lambda } + \left( p^2c^2\frac{m^2}{r^2} + 2\frac{m}{r}p\lambda +  \lambda^2\right)c\frac{m}{r}  } \\
& = \frac{\lambda \sqrt{c^2h^2\left(hp + \lambda\right)^2 + \left(pc^2h^2 + h\lambda  \right)^2 - 2\left(hp + \lambda\right)\left(pc^2h^2 + h\lambda   \right)h }}{\left(pc^2h^2 + h\lambda\right)\sqrt{p^2c^2h^2+\lambda^2 + 2hp\lambda } + \left( p^2c^2h^2 + 2hp\lambda +  \lambda^2\right)ch  } \\
& = \frac{\frac{1}{h^2}\lambda \sqrt{c^2h^2\left(hp + \lambda\right)^2 + \left(pc^2h^2 + h\lambda  \right)^2 - 2\left(hp + \lambda\right)\left(pc^2h^2 + h\lambda   \right)h }}{\frac{1}{h^2}\left(\left(pc^2h^2 + h\lambda\right)\sqrt{p^2c^2h^2+\lambda^2 + 2hp\lambda } + \left( p^2c^2h^2 + 2hp\lambda +  \lambda^2\right)ch \right)  } \\
& = \frac{\lambda}{h}\frac{\sqrt{c^2\left(p + \frac{\lambda}{h}\right)^2 + \left(pc^2 + \frac{\lambda}{h}  \right)^2 - 2\left(p + \frac{\lambda}{h}\right)\left(pc^2 + \frac{\lambda}{h}   \right) }}{\left(pc^2 + \frac{\lambda}{h}\right)\sqrt{p^2c^2+\left(\frac{\lambda}{h}\right)^2 + 2p\frac{\lambda}{h} } + \left( p^2c^2+ 2p\frac{\lambda}{h} +  \left(\frac{\lambda}{h}\right)^2\right)c  } 
\end{split}
\label{eq:final_metric_equation}
\end{equation}

Note that this formula is a function of $\frac{\lambda}{h}$, $p$, and $c$. By expanding this function as a Taylor series, we find 

\begin{equation}
\frac{|\psi'-\phi|}{|\psi'|+|\phi|}  =\frac{\sqrt{{{c}^{2}}-1}\lambda}{2{{c}^{2}}hp}-\frac{\sqrt{{{c}^{2}}-1}\,{{\lambda}^{2}}}{2{{c}^{4}}\,{{h}^{2}}\,{{p}^{2}}}+\mbox{...}
\end{equation}

Noting that $\frac{\sqrt{{{c}^{2}}-1}\,{{\lambda}^{2}}}{2{{c}^{4}}\,{{h}^{2}}\,{{p}^{2}}}$ is always positive and applying Taylor's theorem , we end up with the inequality 
\begin{equation}
\frac{|\psi'-\phi|}{|\psi'|+|\phi|}  \leq \frac{\sqrt{{{c}^{2}}-1}\lambda}{2{{c}^{2}}hp} \leq \frac{1}{4}\frac{\lambda}{hp} 
\label{eq:metic_ineq}
\end{equation}

Where the last step comes from the inequality

\[
\frac{\sqrt{{{c}^{2}}-1}}{{{c}^{2}}} \leq \frac{1}{2}
\]

%% file: main.bbl
\begin{thebibliography}{10}

\bibitem{akyildiz2007survey}
Ian~F Akyildiz, Tommaso Melodia, and Kaushik~R Chowdhury.
\newblock A survey on wireless multimedia sensor networks.
\newblock {\em Computer networks}, 51(4):921--960, 2007.

\bibitem{akyildiz2002survey}
Ian~F Akyildiz, Weilian Su, Yogesh Sankarasubramaniam, and Erdal Cayirci.
\newblock A survey on sensor networks.
\newblock {\em IEEE Communications magazine}, 40(8):102--114, 2002.

\bibitem{arietta2014city}
Sean~M Arietta, Alexei~A Efros, Ravi Ramamoorthi, and Maneesh Agrawala.
\newblock City forensics: Using visual elements to predict non-visual city
  attributes.
\newblock {\em IEEE Transactions on Visualization and Computer Graphics
  (TVCG)}, 20(12):2624--2633, 2014.

\bibitem{basagni2007controlled}
Stefano Basagni, Alessio Carosi, and Chiara Petrioli.
\newblock Controlled vs. uncontrolled mobility in wireless sensor networks:
  Some performance insights.
\newblock In {\em Vehicular Technology Conference, 2007}, pages 269--273,
  Maryland, USA, 2007. IEEE.

\bibitem{behnel2010cython}
S.~Behnel, R.~Bradshaw, C.~Citro, L.~Dalcin, D.S. Seljebotn, and K.~Smith.
\newblock Cython: The best of both worlds.
\newblock {\em Computing in Science Engineering}, 13(2):31 --39, 2011.

\bibitem{bentley1975multidimensional}
Jon~Louis Bentley.
\newblock Multidimensional binary search trees used for associative searching.
\newblock {\em Communications of the ACM}, 18(9):509--517, 1975.

\bibitem{burges1998tutorial}
Christopher~JC Burges.
\newblock A tutorial on support vector machines for pattern recognition.
\newblock {\em Data mining and knowledge discovery}, 2(2):121--167, 1998.

\bibitem{camp2002survey}
Tracy Camp, Jeff Boleng, and Vanessa Davies.
\newblock A survey of mobility models for ad hoc network research.
\newblock {\em Wireless communications and mobile computing}, 2(5):483--502,
  2002.

\bibitem{chapelle2009semi}
Olivier Chapelle, Bernhard Scholkopf, and Alexander Zien.
\newblock Semi-supervised learning (chapelle, o. et al., eds.; 2006)[book
  reviews].
\newblock {\em IEEE Transactions on Neural Networks}, 20(3):542--542, 2009.

\bibitem{consolvo2008activity}
Sunny Consolvo, David~W McDonald, Tammy Toscos, Mike~Y Chen, Jon Froehlich,
  Beverly Harrison, Predrag Klasnja, Anthony LaMarca, Louis LeGrand, Ryan
  Libby, et~al.
\newblock Activity sensing in the wild: a field trial of ubifit garden.
\newblock In {\em Conference on Human Factors in Computing Systems (CHI)},
  pages 1797--1806, Italy, 2008. ACM.

\bibitem{cordts2016cityscapes}
Marius Cordts, Mohamed Omran, Sebastian Ramos, Timo Rehfeld, Markus Enzweiler,
  Rodrigo Benenson, Uwe Franke, Stefan Roth, and Bernt Schiele.
\newblock The cityscapes dataset for semantic urban scene understanding.
\newblock In {\em International Conference on Computer Vision and Pattern
  Recognition (CVPR)}, pages 3213--3223, USA, 2016. IEEE.

\bibitem{dai16rfcn}
Jifeng Dai, Yi~Li, Kaiming He, and Jian Sun.
\newblock {R-FCN}: Object detection via region-based fully convolutional
  networks.
\newblock {\em arXiv preprint arXiv:1605.06409}, 2016.

\bibitem{dalal2005histograms}
Navneet Dalal and Bill Triggs.
\newblock Histograms of oriented gradients for human detection.
\newblock In {\em International Conference on Computer Vision and Pattern
  Recognition (CVPR)}, volume~1, pages 886--893, USA, 2005. IEEE.

\bibitem{davies2000evaluating}
Vanessa~Ann Davies et~al.
\newblock Evaluating mobility models within an ad hoc network.
\newblock Master's thesis, Citeseer, 2000.

\bibitem{everingham2010pascal}
Mark Everingham, Luc Van~Gool, Christopher~KI Williams, John Winn, and Andrew
  Zisserman.
\newblock The pascal visual object classes (voc) challenge.
\newblock {\em International journal of computer vision}, 88(2):303--338, 2010.

\bibitem{gandy2011tensor}
Silvia Gandy, Benjamin Recht, and Isao Yamada.
\newblock Tensor completion and low-n-rank tensor recovery via convex
  optimization.
\newblock {\em Inverse Problems}, 27(2):025010, 2011.

\bibitem{gao2016people}
Chenqiang Gao, Pei Li, Yajun Zhang, Jiang Liu, and Lan Wang.
\newblock People counting based on head detection combining adaboost and cnn in
  crowded surveillance environment.
\newblock {\em Neurocomputing}, 208:108--116, 2016.

\bibitem{geiger2013vision}
Andreas Geiger, Philip Lenz, Christoph Stiller, and Raquel Urtasun.
\newblock Vision meets robotics: The kitti dataset.
\newblock {\em International Journal of Robotics Research (IJRR)}, 2013.

\bibitem{googlemaps}
{Google Inc.}
\newblock (https://maps.google.com), Last accessed March 2017.

\bibitem{hart1968formal}
Peter~E Hart, Nils~J Nilsson, and Bertram Raphael.
\newblock A formal basis for the heuristic determination of minimum cost paths.
\newblock {\em IEEE transactions on Systems Science and Cybernetics},
  4(2):100--107, 1968.

\bibitem{he2016deep}
Kaiming He, Xiangyu Zhang, Shaoqing Ren, and Jian Sun.
\newblock Deep residual learning for image recognition.
\newblock In {\em International Conference on Computer Vision and Pattern
  Recognition (CVPR)}. IEEE, 2016.

\bibitem{helbing1998fluid}
Dirk Helbing.
\newblock A fluid dynamic model for the movement of pedestrians.
\newblock {\em arXiv preprint cond-mat/9805213}, 1998.

\bibitem{huang2005coverage}
Chi-Fu Huang and Yu-Chee Tseng.
\newblock The coverage problem in a wireless sensor network.
\newblock {\em Mobile Networks and Applications}, 10(4):519--528, 2005.

\bibitem{iwata2017estimating}
Tomoharu Iwata, Hitoshi Shimizu, Futoshi Naya, and Naonori Ueda.
\newblock Estimating people flow from spatiotemporal population data via
  collective graphical mixture models.
\newblock {\em ACM Transactions on Spatial Algorithms and Systems (TSAS)},
  3(1):2, 2017.

\bibitem{johnson1996dynamic}
David~B Johnson and David~A Maltz.
\newblock Dynamic source routing in ad hoc wireless networks.
\newblock {\em Mobile computing}, 353(1):153--181, 1996.

\bibitem{karagiorgou2017layered}
Sophia Karagiorgou, Dieter Pfoser, and Dimitrios Skoutas.
\newblock A layered approach for more robust generation of road network maps
  from vehicle tracking data.
\newblock {\em ACM Transactions on Spatial Algorithms and Systems (TSAS)},
  3(1):3, 2017.

\bibitem{kleinrock1976queueing}
Leonard Kleinrock.
\newblock {\em Queueing systems, volume 2: Computer applications}, volume~66.
\newblock John Wiley \& Sons, USA, 1976.

\bibitem{knuth1997art}
Donald~Ervin Knuth.
\newblock {\em The art of computer programming}, volume~3.
\newblock Addison-Wesley, USA, 1997.

\bibitem{krizhevsky2009learning}
Alex Krizhevsky et~al.
\newblock Learning multiple layers of features from tiny images.
\newblock Technical report, Citeseer, 2009.

\bibitem{krizhevsky2012imagenet}
Alex Krizhevsky, Ilya Sutskever, and Geoffrey~E Hinton.
\newblock Imagenet classification with deep convolutional neural networks.
\newblock In {\em Advances in neural information processing systems}, pages
  1097--1105, Nevada, USA, 2012.

\bibitem{lane2010survey}
Nicholas~D Lane, Emiliano Miluzzo, Hong Lu, Daniel Peebles, Tanzeem Choudhury,
  and Andrew~T Campbell.
\newblock A survey of mobile phone sensing.
\newblock {\em IEEE Communications magazine}, 48(9):140--150, 2010.

\bibitem{law2007simulation}
Averill~M Law, W~David Kelton, and W~David Kelton.
\newblock {\em Simulation modeling and analysis}, volume~3.
\newblock McGraw-Hill New York, Arizona, USA, 2007.

\bibitem{lee2009dissemination}
Uichin Lee, Eugenio Magistretti, Mario Gerla, Paolo Bellavista, and Antonio
  Corradi.
\newblock Dissemination and harvesting of urban data using vehicular sensing
  platforms.
\newblock {\em IEEE Transactions on Vehicular Technology}, 58(2):882--901,
  2009.

\bibitem{lesser2012distributed}
Victor Lesser, Charles~L Ortiz~Jr, and Milind Tambe.
\newblock {\em Distributed sensor networks: A multiagent perspective},
  volume~9.
\newblock Springer Science \& Business Media, 2012.

\bibitem{li2017city}
Weizi Li, David Wolinski, and Ming~C Lin.
\newblock City-scale traffic animation using statistical learning and
  metamodel-based optimization.
\newblock {\em ACM Transactions on Graphics (TOG)}, 36(6):200, 2017.

\bibitem{liang1999predictive}
Ben Liang and Zygmunt~J Haas.
\newblock Predictive distance-based mobility management for pcs networks.
\newblock In {\em INFOCOM'99. Eighteenth Annual Joint Conference of the IEEE
  Computer and Communications Societies. Proceedings. IEEE}, volume~3, pages
  1377--1384, New York, USA, 1999. IEEE, IEEE.

\bibitem{lin2014microsoft}
Tsung-Yi Lin, Michael Maire, Serge Belongie, James Hays, Pietro Perona, Deva
  Ramanan, Piotr Doll{\'a}r, and C~Lawrence Zitnick.
\newblock Microsoft coco: Common objects in context.
\newblock In {\em Computer Vision--ECCV 2014}, pages 740--755. Springer,
  Zurich, Switzerland, 2014.

\bibitem{maddern20171}
Will Maddern, Geoffrey Pascoe, Chris Linegar, and Paul Newman.
\newblock 1 year, 1000 km: The oxford robotcar dataset.
\newblock {\em The International Journal of Robotics Research}, 36(1):3--15,
  2017.

\bibitem{matsumoto1998mersenne}
Makoto Matsumoto and Takuji Nishimura.
\newblock Mersenne twister: a 623-dimensionally equidistributed uniform
  pseudo-random number generator.
\newblock {\em ACM Transactions on Modeling and Computer Simulation (TOMACS)},
  8(1):3--30, 1998.

\bibitem{miranda2018shadow}
Fabio Miranda, Harish Doraiswamy, Marcos Lage, Luc Wilson, Mondrian Hsieh, and
  Claudio~T Silva.
\newblock Shadow accrual maps: Efficient accumulation of city-scale shadows
  over time.
\newblock {\em IEEE Transactions on Visualization and Computer Graphics}, 2018.

\bibitem{niazi2011novel}
Muaz~A Niazi and Amir Hussain.
\newblock A novel agent-based simulation framework for sensing in complex
  adaptive environments.
\newblock {\em IEEE Sensors Journal}, 11(2):404--412, 2011.

\bibitem{nycopendata}
{NYC open data}.
\newblock (https://opendata.cityofnewyork.us/), Last accessed March 2017.

\bibitem{oh2011large}
Sangmin Oh, Anthony Hoogs, Amitha Perera, Naresh Cuntoor, Chia-Chih Chen,
  Jong~Taek Lee, Saurajit Mukherjee, JK~Aggarwal, Hyungtae Lee, Larry Davis,
  et~al.
\newblock A large-scale benchmark dataset for event recognition in surveillance
  video.
\newblock In {\em International Conference on Computer Vision and Pattern
  Recognition (CVPR)}, pages 3153--3160, Colorado, USA, 2011. IEEE.

\bibitem{openstreetmap}
{OpenStreetMap}.
\newblock {Planet dump retrieved from https://planet.osm.org }.
\newblock \url{ https://www.openstreetmap.org }, 2017.

\bibitem{othman2012wireless}
Mohd~Fauzi Othman and Khairunnisa Shazali.
\newblock Wireless sensor network applications: A study in environment
  monitoring system.
\newblock {\em Procedia Engineering}, 41:1204--1210, 2012.

\bibitem{park1988random}
Stephen~K. Park and Keith~W. Miller.
\newblock Random number generators: good ones are hard to find.
\newblock {\em Communications of the ACM}, 31(10):1192--1201, 1988.

\bibitem{photosynth}
Photosynth.
\newblock (https://blogs.msdn.microsoft.com/photosynth/
  2017/02/06/microsoft-photosynth-has-been-shut-down/), Last accessed March
  2017.

\bibitem{rana2010ear}
Rajib~Kumar Rana, Chun~Tung Chou, Salil~S Kanhere, Nirupama Bulusu, and Wen Hu.
\newblock Ear-phone: an end-to-end participatory urban noise mapping system.
\newblock In {\em Proceedings of the 9th ACM/IEEE International Conference on
  Information Processing in Sensor Networks}, pages 105--116. ACM, 2010.

\bibitem{reades2007cellular}
Jonathan Reades, Francesco Calabrese, Andres Sevtsuk, and Carlo Ratti.
\newblock Cellular census: Explorations in urban data collection.
\newblock {\em IEEE Pervasive computing}, 6(3):30--38, 2007.

\bibitem{ren2015faster}
Shaoqing Ren, Kaiming He, Ross Girshick, and Jian Sun.
\newblock Faster r-cnn: Towards real-time object detection with region proposal
  networks.
\newblock In {\em Advances in Neural Information Processing Systems}, pages
  91--99, 2015.

\bibitem{russakovsky2015imagenet}
Olga Russakovsky, Jia Deng, Hao Su, Jonathan Krause, Sanjeev Satheesh, Sean Ma,
  Zhiheng Huang, Andrej Karpathy, Aditya Khosla, Michael Bernstein, et~al.
\newblock Imagenet large scale visual recognition challenge.
\newblock {\em International Journal of Computer Vision}, 115(3):211--252,
  2015.

\bibitem{saqib2018person}
Muhammad Saqib, Sultan~Daud Khan, Nabin Sharma, and Michael Blumenstein.
\newblock Person head detection in multiple scales using deep convolutional
  neural networks.
\newblock In {\em 2018 International Joint Conference on Neural Networks
  (IJCNN)}, pages 1--7. IEEE, 2018.

\bibitem{sheng2012energy}
Xiang Sheng, Jian Tang, and Weiyi Zhang.
\newblock Energy-efficient collaborative sensing with mobile phones.
\newblock In {\em INFOCOM, 2012 Proceedings IEEE}, pages 1916--1924, Florida,
  USA, 2012. IEEE.

\bibitem{shi2009automatic}
Wenhuan Shi, Shuhan Shen, and Yuncai Liu.
\newblock Automatic generation of road network map from massive gps, vehicle
  trajectories.
\newblock In {\em Intelligent Transportation Systems, 2009. ITSC'09. 12th
  International IEEE Conference on}, pages 1--6, Missouri, USA, 2009. IEEE.

\bibitem{silva2019integrated}
Claudio~T. Silva, Juliana Freire, Fabio Miranda, Marcos Lage, Harish
  Doraiswamy, Maryam Hosseini, Eric~K. Tokuda, Gabriel Ferreira, and Roberto~M.
  Cesar~Jr.
\newblock Integrated analytics and visualization for multi-modality
  transportation data.
\newblock Technical report, C2SMART Connected Cities for Smart Transportation,
  2019.

\bibitem{szegedy2015going}
Christian Szegedy, Wei Liu, Yangqing Jia, Pierre Sermanet, Scott Reed, Dragomir
  Anguelov, Dumitru Erhan, Vincent Vanhoucke, Andrew Rabinovich, et~al.
\newblock Going deeper with convolutions.
\newblock In {\em International Conference on Computer Vision and Pattern
  Recognition (CVPR)}, Massachusetts, USA, 2015.

\bibitem{tian2002graph}
Jing Tian, Jorg Hahner, Christian Becker, Illya Stepanov, and Kurt Rothermel.
\newblock Graph-based mobility model for mobile ad hoc network simulation.
\newblock In {\em Simulation Symposium, 2002. Proceedings. 35th Annual}, pages
  337--344, California, USA, 2002. IEEE, IEEE.

\bibitem{titzer2005avrora}
Ben~L Titzer, Daniel~K Lee, and Jens Palsberg.
\newblock Avrora: Scalable sensor network simulation with precise timing.
\newblock In {\em International Symposium on Information Processing in Sensor
  Networks}, pages 477--482, Tennessee, USA, 2005. IEEE.

\bibitem{tokuda2019quantifying}
Eric~K. Tokuda, Roberto~M. Cesar, and Claudio~T. Silva.
\newblock Quantifying the presence of graffiti in urban environments.
\newblock In {\em International Conference on Big Data and Smart Computing
  (BigComp)}, pages 1--4, Japan, 2019. IEEE.

\bibitem{tokuda2018a}
Eric~K. Tokuda, Gabriel B.~A. Ferreira, Claudio Silva, and Roberto~M. Cesar-Jr.
\newblock A novel semi-supervised detection approach with weak annotation.
\newblock In {\em Image Analysis and Interpretation (SSIAI), 2018 IEEE
  Southwest Symposium on}, Nevada, USA, 2018. IEEE.

\bibitem{united2016air}
United States Environment Protection Agency Air~Data USEEPAAD.
\newblock (https://www3.epa.gov/airdata/ad\_data\_daily.html), Last accessed
  March 2017.

\bibitem{vanegas2012automatic}
Carlos~A Vanegas, Daniel~G Aliaga, and Bedrich Benes.
\newblock Automatic extraction of manhattan-world building masses from 3d laser
  range scans.
\newblock {\em Visualization and Computer Graphics, IEEE Transactions on},
  18(10):1627--1637, 2012.

\bibitem{vezzani2010video}
Roberto Vezzani and Rita Cucchiara.
\newblock Video surveillance online repository (visor): an integrated
  framework.
\newblock {\em Multimedia Tools and Applications}, 50(2):359--380, 2010.

\bibitem{vinyals2011survey}
Meritxell Vinyals, Juan~A Rodriguez-Aguilar, and Jesus Cerquides.
\newblock A survey on sensor networks from a multiagent perspective.
\newblock {\em The Computer Journal}, 54(3):455--470, 2011.

\bibitem{wang2003bidding}
Guiling Wang, Guohong Cao, and Tom LaPorta.
\newblock A bidding protocol for deploying mobile sensors.
\newblock In {\em International Conference on Network Protocols}, pages
  315--324, Georgia, USA, 2003. IEEE.

\bibitem{whyte2012city}
William~H Whyte.
\newblock {\em City: Rediscovering the center}.
\newblock University of Pennsylvania Press, Pennsylvania, USA, 2012.

\bibitem{yang2003counting}
Danny~B Yang, Leonidas~J Guibas, et~al.
\newblock Counting people in crowds with a real-time network of simple image
  sensors.
\newblock In {\em International Conference on Computer Vision Workshops}, page
  122, Nice, France, 2003. IEEE.

\bibitem{zemene2018large}
Eyasu Zemene, Yonatan~Tariku Tesfaye, Haroon Idrees, Andrea Prati, Marcello
  Pelillo, and Mubarak Shah.
\newblock Large-scale image geo-localization using dominant sets.
\newblock {\em IEEE transactions on pattern analysis and machine intelligence},
  41(1):148--161, 2018.

\bibitem{zheng2014urban}
Yu~Zheng, Licia Capra, Ouri Wolfson, and Hai Yang.
\newblock Urban computing: concepts, methodologies, and applications.
\newblock {\em ACM Transactions on Intelligent Systems and Technology (TIST)},
  5(3):38, 2014.

\bibitem{zhu2005semi}
Xiaojin~Jerry Zhu.
\newblock Semi-supervised learning literature survey.
\newblock Technical report, University of Wisconsin-Madison Department of
  Computer Sciences, 2005.

\end{thebibliography}
